\documentclass{article} 
\usepackage{iclr2027_conference,times}

\usepackage{amsmath,amsfonts,bm}

\def\eqref#1{equation~\ref{#1}}

\def\1{\bm{1}}

\DeclareMathAlphabet{\mathsfit}{\encodingdefault}{\sfdefault}{m}{sl}
\SetMathAlphabet{\mathsfit}{bold}{\encodingdefault}{\sfdefault}{bx}{n}

\usepackage{hyperref}
\usepackage{url}

\usepackage{graphicx}
\usepackage{booktabs} 
\usepackage[table]{xcolor}
\usepackage{multirow}

\title{\textbf{RESETTLE}: Robotic Recovery through Disagreement-Triggered
Retrieval and Efficient Corrective Control}

\author{%
\begin{minipage}[t]{\dimexpr\textwidth-2\tabcolsep\relax}
\centering\normalfont
\textbf{%
\makebox[\linewidth][s]{%
Yuxin Chen\textsuperscript{1,*}\hfill
Senqiao Yang\textsuperscript{2,*}\hfill
Zixuan Wang\textsuperscript{1}\hfill
Jinhui Ye\textsuperscript{1}}\\[0.25em]
\makebox[\linewidth][s]{%
Changsheng Lu\textsuperscript{1}\hfill
Pengguang Chen\textsuperscript{3}\hfill
Shu Liu\textsuperscript{3}\hfill
Zhuotao Tian\textsuperscript{4,5}\hfill
Jiaya Jia\textsuperscript{1}}}\\[0.65em]
\fontsize{8.5}{11}\selectfont
\mbox{\textsuperscript{1}The Hong Kong University of Science and Technology\quad
\textsuperscript{2}The Chinese University of Hong Kong}\\[0.15em]
\mbox{\textsuperscript{3}SmartMore\quad
\textsuperscript{4}Harbin Institute of Technology, Shenzhen\quad
\textsuperscript{5}Shenzhen Loop Area Institute}
\end{minipage}%
}

\iclrfinalcopy
\begin{document}

\maketitle
\fancyhead{}
\renewcommand{\headrulewidth}{0pt}
\begingroup
\renewcommand{\thefootnote}{\fnsymbol{footnote}}
\footnotetext[1]{Equal contribution.}
\endgroup

\begin{abstract}
Reliable robotic manipulation requires timely intervention to correct emerging deviations and restore progress after execution errors. However, recovery methods based on repeated vision-language reasoning or iterative online optimization can incur substantial latency, delaying intervention.
To address these challenges, we introduce \textbf{RESETTLE} (\textbf{R}obotic r\textbf{E}covery through di\textbf{S}agr\textbf{E}ement-\textbf{T}riggered re\textbf{T}rieva\textbf{L} and \textbf{E}fficient Corrective Control), a model-agnostic framework that provides computationally efficient recovery at the action-execution interface of frozen robot policies.
RESETTLE triggers recovery when two action proposals independently sampled under identical conditioning persistently disagree. It retrieves a same-task demonstration reference using an adapted V-JEPA encoder and combines a state-servo prior with a guarded visual residual to execute one corrective action without online trajectory optimization or additional vision-language reasoning, then returns control to the base policy.
Across six base policies in simulation, RESETTLE achieves up to 8.70\%, 6.28\%, and 6.83\% absolute success-rate gains on LIBERO-Plus, Meta-World, and RoboCasa Tabletop, respectively, with further improvements on four real-world tasks using two policies. In QwenPI-based comparisons, its monitoring-and-recovery computation latency is 74.04\%--93.57\% lower than VoLoAgent's monitoring-and-planning latency for grasp and place tool calls. It also raises Harness VLA's LIBERO-Pro Swap success from 42\% to 50\%, demonstrating compatibility with high-level agentic planning. 
Code available at: \url{https://github.com/JIA-Lab-research/RESETTLE}
\end{abstract}

\section{Introduction}
Timely recovery is essential for reliable robotic manipulation, helping robots correct potentially harmful motions before an error occurs and restore task progress after errors. For example, adjusting the gripper orientation while approaching a moka pot can help avoid an ineffective grasp, whereas repositioning the gripper after a bowl is dropped can enable regrasping and subsequent placement.
These interventions support continued execution by addressing both emerging problems and the consequences of earlier mistakes. Despite substantial progress in vision--language--action (VLA) policies and world--action models (WAMs)~\citep{intelligence2025pi_,ye2026starvla,yang2026beyond,kim2024openvla,yuan2026fast,ye2026world,cen2025worldvla,ma2026dit4dit}, environmental changes and imperfect actions can still drive execution into unfamiliar states where the policy struggles to make progress, highlighting the need for timely corrective intervention.

\begin{figure}[!t]
    \centering
    \includegraphics[width=\linewidth]{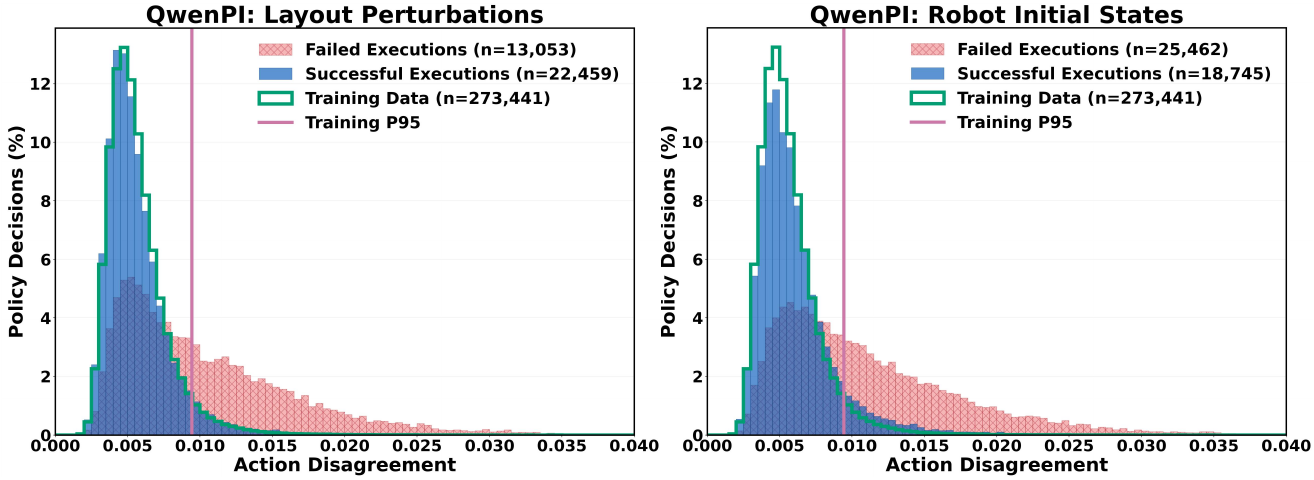}
    \caption{QwenPI action disagreement on LIBERO-Plus for perturbations in layout \textit{(left)} and robot-initialization \textit{(right)}. Histograms show within-group decision percentages. ``Training Data'' denotes training-data inference; ``Training P95'' marks its 95th percentile. Failures exhibit heavier tails.}
    \label{fig:analysis_action}
    \vspace{-6pt}
\end{figure}


To equip robots with such corrective capabilities,
existing approaches broadly follow three strategies.
\textbf{First}, corrective policy training incorporates
recovery behavior through additional supervision:
DAgger collects expert labels on learner-visited
states~\citep{ross2011reduction}, while IntervenGen
generates recovery trajectories from human
interventions~\citep{hoque2024intervengen}.
Both expand the policy's corrective experience but require
additional data and policy updates.
\textbf{Second}, VLM-supervised recovery coordinates
corrective behaviors through semantic assessment.
Agentic Robot uses visual verification to trigger
corrections and retries~\citep{yang2025agentic},
while VoLoAgent switches between policy execution and
recovery tools~\citep{chen2026volo}.
This supports flexible task-level recovery, but repeated
verification and recovery computation introduce latency.
\textbf{Third}, inference-time action correction preserves
base-policy weights while adjusting action generation. LPB uses expert-feature distance to trigger dynamics-guided denoising on diffusion policies~\citep{sun2026latent}. When extended to generalized VLA/WAMs, however, we observe greater overlap between feature-distance distributions of successful and failed executions, reducing its effectiveness for error detection; VLA-Corrector detects visual prediction discrepancies and applies gradient-guided replanning~\citep{pan2026vla}, 
while the online gradient computation adds recovery overhead. 
To push forward this line of research direction, we ask:
\emph{Can we efficiently identify when a frozen policy
needs assistance and directly generate corrections
before deviations escalate into task failure?}


To answer this question, we introduce \textbf{RESETTLE}, a framework for efficient action-level
recovery with frozen robot policies.
Many generative robot policies sample actions from noise
conditioned on observations and task
instructions~\citep{chi2025diffusion,black2024pi_0}.
This sampling process allows us to probe the stability
of action choices under identical conditioning.
For QwenPI, Figure~\ref{fig:analysis_action} shows that
training-data inference and successful executions align
in the dominant low-disagreement region, whereas failed
executions exhibit heavier tails.
Motivated by this observation, RESETTLE monitors two
independently sampled proposals and triggers intervention
when their disagreement persists.
To guide recovery, we use successful demonstrations as
empirical anchors without requiring them to cover the
policy's entire feasible operating range.
We initialize the recovery representation from
V-JEPA 2-AC~\citep{assran2025v} and adapt it through
dynamics prediction and corrective-action supervision,
encouraging features useful for reference matching
and control.
The adapted encoder retrieves the nearest same-task
demonstration reference in its representation space,
together with the associated robot state.
A state-servo prior provides motion toward this reference
state, while a guarded visual residual uses current and
reference observations to adjust the motion to the scene.
Each intervention executes one corrective action before
returning control to the base policy with a fresh observation.
This design directly constructs recovery actions without
online trajectory optimization or additional
vision-language reasoning.

Our experiments assess both the effectiveness and
computational cost of local recovery.
Across six base policies on LIBERO-Plus, Meta-World,
and RoboCasa Tabletop~\citep{fei2026libero,yu2020meta,nasiriany2024robocasa},
RESETTLE improves aggregate task success while keeping
the base policies frozen.
Further improvements across four real-world tasks with
two policies support its applicability beyond simulation.
In QwenPI-based comparisons, measured monitoring-and-recovery computation latency is \textbf{74.04\%--93.57\%} lower than VoLoAgent's combined monitoring-and-planning latency for grasp and place tool calls, supporting the efficiency of direct correction.
Integration with Harness VLA on LIBERO-Pro
Swap~\citep{zhang2026harness,zhou2025libero} further
demonstrates that action-level recovery complements
high-level agentic planning.

Our contributions are twofold:
\textbf{i)} We introduce \textbf{RESETTLE}, an efficient
action-level recovery framework that combines persistent
disagreement monitoring, demonstration retrieval, and direct
corrective control for frozen VLA policies and WAMs,
without base-policy fine-tuning, additional vision-language
reasoning calls for recovery, or online trajectory optimization;
\textbf{ii)} We validate its effectiveness in simulation
and real-world manipulation, quantify reductions in
monitoring-and-recovery computation latency, and demonstrate
that local recovery complements high-level agentic planning
through integration with Harness VLA.

\section{Related Work}
\label{related_work}

\textbf{Generative VLA and World-Action Models.}
Generative policies such as Diffusion Policy, $\pi_0$,
and SmolVLA produce actions through diffusion or
flow matching~\citep{chi2025diffusion,black2024pi_0,shukor2025smolvla}.
Video-action architectures such as DiT4DiT and FastWAM
incorporate visual dynamics into policy learning,
with FastWAM omitting explicit future-video generation
at inference~\citep{ma2026dit4dit,yuan2026fast}.
Together, these approaches advance policy representations
and action generation.
RESETTLE complements these capabilities through
execution recovery at the action interface, leaving
base-policy weights and action-generation mechanisms
unchanged.

\textbf{Latent World Models and JEPA-Based Control.}
DINO-WM and V-JEPA 2-AC predict visual latents under candidate actions and use online optimization for goal-directed control~\citep{zhou2024dino,assran2025v}. INTACT learns an intent-to-action interface with latent dynamics supervision, enabling control without online search~\citep{sun2026intact}. RESETTLE uses dynamics supervision during training, initializing its encoder and predictor from V-JEPA-2-AC. Its deployment role is selective recovery: demonstration states anchor a servo prior, refined by a guarded visual residual, before control returns to a frozen policy. The dynamics predictor is inactive at deployment.

\textbf{Failure Recovery in Visuomotor Policies.}
DAgger incorporates expert labels on learner-visited states
into policy training, while RACER uses language-annotated
recovery trajectories~\citep{ross2011reduction,dai2025racer}.
Further, \cite{agia2024unpacking} combine temporal action consistency with VLM for progress monitoring. The existing works show that semantic supervision supports failure diagnosis~\citep{duan2024aha}, constraint-based replanning~\citep{guo2024doremi}, and recovery coordination 
~\citep{yang2025agentic,chen2026volo}. 
LPB and VLA-Corrector preserve policy weights, but they rely on gradient-guided action
refinement which adds computational burden~\citep{sun2026latent,pan2026vla}. Instead, our RESETTLE monitors persistent disagreement under identical
conditioning and directly constructs reference-based corrections, avoiding online dynamics prediction,
gradient guidance, and additional VLM reasoning while
complementing high-level supervision.

\section{Preliminaries}
\label{sec:preliminaries}

\paragraph{Diffusion and Flow Matching in VLAs and WAMs.}
We consider VLA and WAM policies that generate action chunks
via diffusion or flow matching, conditioned on observations
and task instructions.
Each chunk specifies commands for multiple future control steps.
Diffusion models learn to denoise action sequences,
while flow-matching models learn a velocity field that
transforms noise into actions~\citep{chi2025diffusion,black2024pi_0}. At inference, sampled noise is iteratively transformed into an action chunk. Different noise samples can yield different chunks
under identical conditioning, reflecting variability in the
learned conditional action distribution. Comparing these
samples provides an action-disagreement signal without
updating the base policy's weights.

\paragraph{Action-Conditioned Prediction with V-JEPA 2.}
V-JEPA 2 learns visual representations from large-scale unlabeled
video data by predicting masked video regions in a learned feature space.
Its action-conditioned extension, V-JEPA 2-AC, predicts future
visual representations from visual context, robot states,
and actions~\citep{assran2025v}.
This formulation motivates learning representations that capture
action consequences, providing a foundation for selecting
corrective actions during recovery.
We adapt the pretrained V-JEPA 2 encoder offline using
action-conditioned prediction with residual supervision.
At deployment, the adapted encoder supports reference retrieval
and residual action prediction to guide corrective actions.
\section{RESETTLE: Proposed Robotic Recovery Method}
\label{sec:RESETTLE}



\begin{figure}[!t]
\centering
\includegraphics[width=\linewidth]{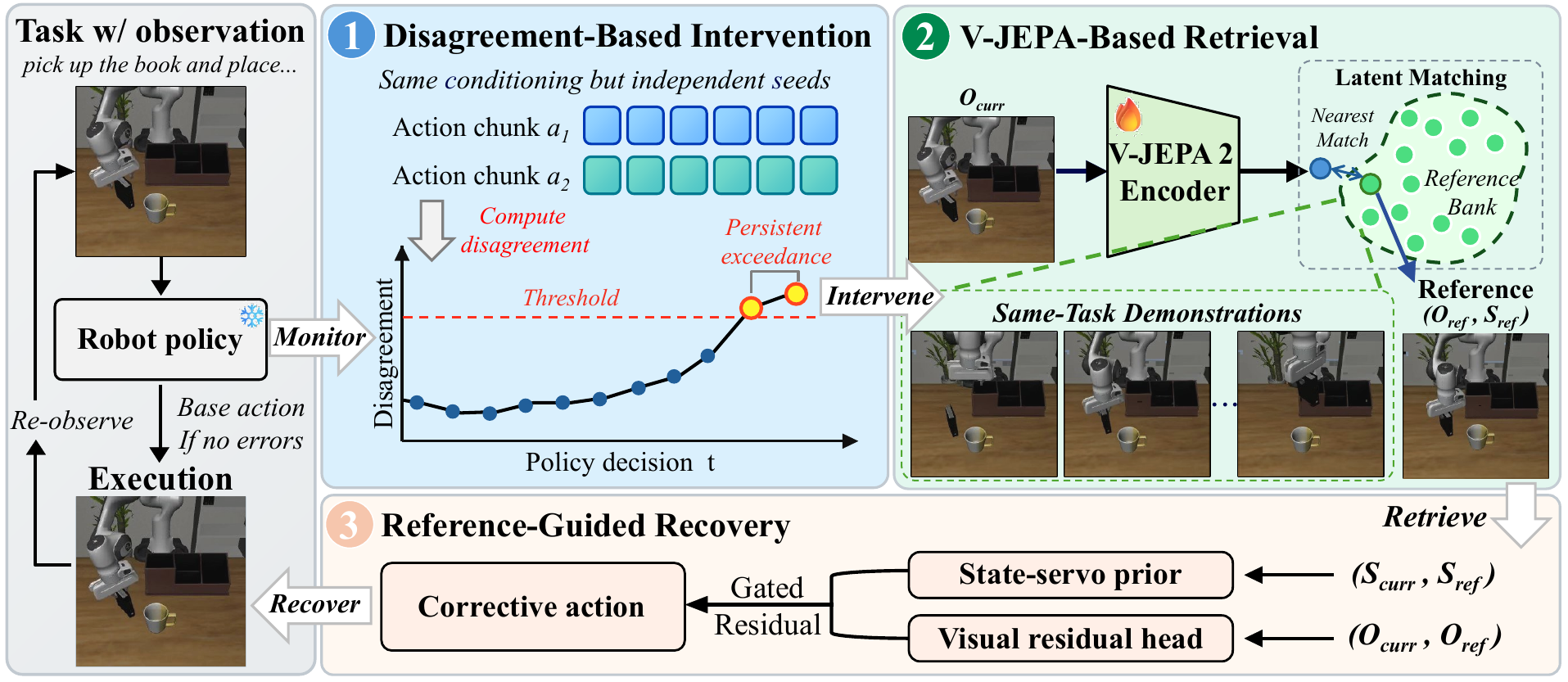}
\caption{
\textbf{Overview of RESETTLE.}
Disagreement-Based Intervention module detects the execution error and initiates V-JEPA-Based
Retrieval, followed by Reference-Guided Recovery using
a state-servo prior and a guarded visual residual. The $a_1, a_2$ are action chunks sampled with independent seeds;
$O_{\mathrm{curr/ref}}$ and $S_{\mathrm{curr/ref}}$ denote
current/reference observations and measured robot states.
Each intervention executes one native control tick before
re-observation and return to the base policy. `Snowflake' signifies frozen policy while `flame' refers to offline encoder adaptation.
}
\label{fig:method}
\end{figure}

RESETTLE augments a frozen robot policy with recovery at
its action-execution interface. As shown in Figure~\ref{fig:method}, our method addresses three decisions: \textbf{\textit{1) when to intervene, 2) which reference to retrieve,
and 3) how to recover}}. Given the current observation, robot
state, and task identity, it monitors action disagreement,
retrieves a successful reference when intervention is
triggered, and generates a corrective action efficiently. Note that our method requires no access to the base policy's internal representations.

\paragraph{Disagreement-Based Intervention.}
At each nominal decision, we generate two action chunks from
independent noise samples under identical observations and
task conditioning. We measure their temporally weighted
disagreement using an embodiment-specific action metric.
Recovery is triggered when disagreement exceeds a calibrated
threshold at two consecutive decisions, suppressing responses
to isolated sampling fluctuations.
For each base policy, thresholds are calibrated as the 95th percentile (P95) of disagreement scores from
offline inference on training data, except in the threshold-percentile ablation.
For each base policy, LIBERO-Plus uses a single training-derived
P95 threshold shared across all perturbation variants and tasks.
The other benchmarks use a benchmark-wide threshold shared across tasks.
Without a trigger, the first proposal follows nominal
execution. After intervention, the exceedance history resets.
Disagreement signals unstable action selection rather than
directly testing physical feasibility or task correctness.

\paragraph{V-JEPA-Based Retrieval.}
When intervention is triggered, the adapted V-JEPA-2 encoder
maps the current observation into a pooled, normalized
representation. We retrieve the nearest representation by
cosine distance from successful training demonstrations of
the same task. Matching seeks a reference relevant to the
current execution phase. Each retrieved frame supplies both
its visual observation and the associated robot state for
corrective control.

\paragraph{Reference-Guided Recovery.}
Successful demonstrations provide empirical references for
corrective control independently of the current, potentially
unreliable policy proposal. These references guide local
recovery without defining all feasible states or guaranteeing
a feasible recovery path. Displacement from the current to the reference robot state
defines a state-servo prior in native control coordinates.
Current and reference observations provide scene context,
including object layout and contact conditions, for a learned
visual residual that adjusts this prior.
The residual also conditions on robot state, previous-action
context, and the prior, adapting the correction to the
current configuration. The guard requires sufficient alignment with the prior
and limits the residual's relative magnitude.
Accepted residuals are added to the prior before clipping
to native command limits; rejection preserves the bounded
servo action.
Each intervention discards pending base-policy actions,
executes one native control action, and returns control
to the base policy with a fresh observation.

\paragraph{Training and deployment.}
We initialize the recovery encoder and auxiliary dynamics
predictor from V-JEPA-2-AC~\citep{assran2025v} and jointly
train them with an embodiment-specific residual head on
expert demonstrations.
A future frame from the same demonstration supplies the
training reference, and the residual head learns to adjust
the resulting servo prior to match the demonstrated motion.
Training combines residual-action supervision, native-action
alignment, latent dynamics prediction, and guard and
saturation penalties; the base policy remains frozen.
At deployment, the encoder supports retrieval and residual
prediction for corrective action generation, while the
dynamics predictor is inactive.
Recovery requires neither online trajectory optimization
nor additional LLM/VLM reasoning calls and can also operate
beneath a high-level planner.
Appendix~\ref{app:resettle_method} collects the mathematical
formulation and implementation details.

\section{Experiments}

\subsection{Experimental Setup}

\begin{figure}[!t]
    \centering
    \includegraphics[width=\linewidth]{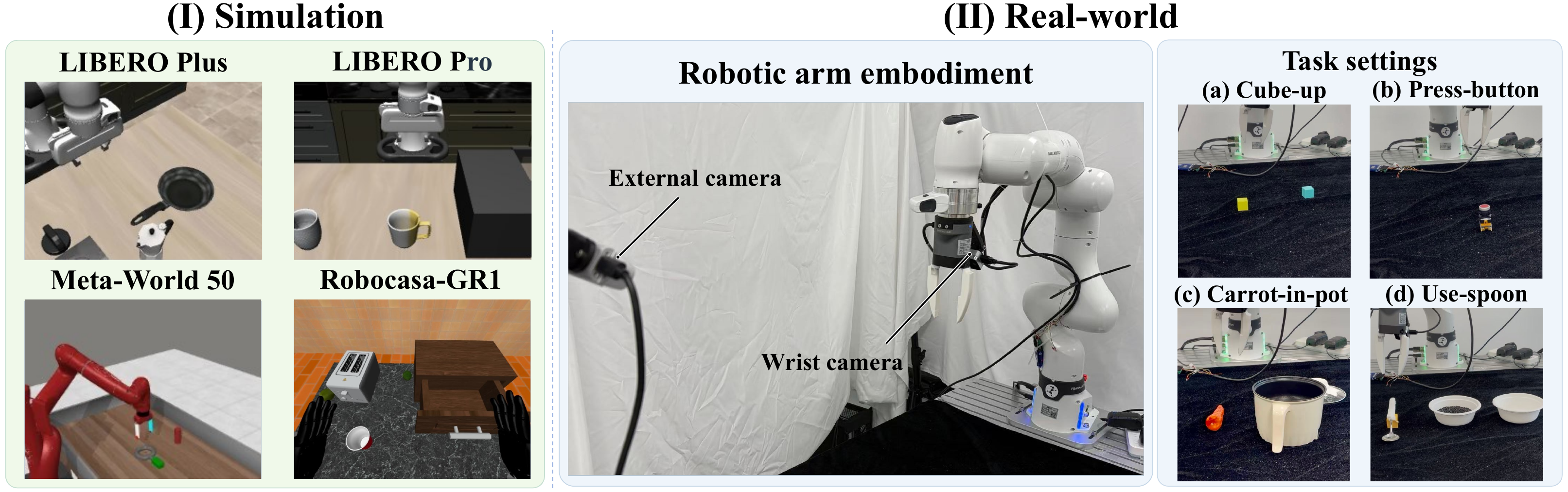}
    \caption{Experimental settings across four simulation benchmarks
    and four real-world manipulation tasks using a Franka Research 3
    robot with external and wrist-mounted cameras.}
    \label{fig:exp_setting}
\end{figure}

We evaluate RESETTLE in four simulation settings and on
four real-world manipulation tasks to assess execution
recovery and compatibility with different policies and
hierarchical agents.
The simulation evaluation includes FastWAM, $\pi_{0.5}$,
and QwenPI on LIBERO-Plus; QwenPI and SmolVLA on
Meta-World; LDA and QwenGR00T on RoboCasa Tabletop;
and $\pi_{0.5}$ with and without a GPT-5.5 planner on
LIBERO-Pro Swap.
Figure~\ref{fig:exp_setting} illustrates the experimental
settings.
Local policy and recovery inference runs on NVIDIA
RTX 4090D GPUs.

\begin{table}[!t]
\centering
\caption{\textbf{Results on LIBERO-Plus.}
Per-dimension success rates (\%) on the LIBERO-Plus robustness benchmark.
Cam.: camera; Lang.: language; BG: background; Robot: robot initialization.
All models are trained on standard LIBERO and evaluated on the perturbed
test set. Average aggregates all evaluation cases, weighting each dimension
by its number of cases. ``Ours'' denotes RESETTLE, while
Ours$^{\dagger}$ denotes RESETTLE without the residual module. 
}
\label{tab:libero_plus_robust}
\vspace{5pt}

\setlength{\tabcolsep}{1.6pt}
\renewcommand{\arraystretch}{1.12}

\begin{tabular}{@{}lcccccccc@{}}
\toprule
Model
& Cam. & Robot & Lang. & Light
& BG & Noise & Layout & Average \\
\midrule

OpenVLA
& 0.8 & 3.5 & 23.0 & 8.1
& 34.8 & 15.2 & 28.5 & 15.6 \\

OpenVLA-OFT
& 56.4 & 31.9 & 79.5 & 88.7
& 93.3 & 75.8 & 74.2 & 69.6 \\

NORA
& 2.2 & 37.0 & 65.1 & 45.7
& 58.6 & 12.8 & 62.1 & 39.0 \\

WorldVLA
& 0.1 & 27.9 & 41.6 & 43.7
& 17.1 & 10.9 & 38.0 & 25.0 \\

UniVLA
& 1.8 & 46.2 & 69.6 & 69.0
& 81.0 & 21.2 & 31.9 & 42.9 \\

$\pi_{0}$
& 13.8 & 6.0 & 58.8 & 85.0
& 81.4 & 79.0 & 68.8 & 53.6 \\

$\pi_{0}$-FAST
& 65.1 & 21.6 & 61.0 & 73.2
& 73.2 & 74.4 & 68.8 & 61.6 \\

RIPT-VLA
& 55.2 & 31.2 & 77.5 & 88.3
& 91.6 & 73.5 & 74.2 & 68.4 \\

Abot-M0
& 60.4 & 67.9 & 86.4 & 96.2
& 91.6 & 86.4 & 82.6 & 80.5 \\

Qwen3VL-OFT
& 47.0 & 60.1 & 87.0 & 96.3
& 95.3 & 73.1 & 79.2 & 75.0 \\

VLAct
& 73.9 & 68.4 & 81.5 & 96.7
& 96.7 & 86.0 & 83.3 & 82.6 \\

\midrule

FastWAM / Base.
& 15.9
& 44.4
& 66.4
& 79.3
& 51.9
& 36.9
& 61.1
& 49.3 \\

\rowcolor{green!4}
FastWAM / Ours$^{\dagger}$
& 24.0\,{\scriptsize\textcolor{green!45!black}{\ensuremath{8.1\!\uparrow}}}
& 56.5\,{\scriptsize\textcolor{green!45!black}{\ensuremath{12.1\!\uparrow}}}
& 73.9\,{\scriptsize\textcolor{green!45!black}{\ensuremath{7.5\!\uparrow}}}
& 86.5\,{\scriptsize\textcolor{green!45!black}{\ensuremath{7.2\!\uparrow}}}
& 56.6\,{\scriptsize\textcolor{green!45!black}{\ensuremath{4.7\!\uparrow}}}
& 38.9\,{\scriptsize\textcolor{green!45!black}{\ensuremath{2.0\!\uparrow}}}
& 64.7\,{\scriptsize\textcolor{green!45!black}{\ensuremath{3.6\!\uparrow}}}
& 55.9\,{\scriptsize\textcolor{green!45!black}{\ensuremath{6.6\!\uparrow}}} \\

\rowcolor{green!8}
FastWAM / \textbf{Ours}
& \textbf{25.0}\,{\scriptsize\textcolor{green!45!black}{\ensuremath{9.1\!\uparrow}}}
& \textbf{61.1}\,{\scriptsize\textcolor{green!45!black}{\ensuremath{16.7\!\uparrow}}}
& \textbf{75.7}\,{\scriptsize\textcolor{green!45!black}{\ensuremath{9.3\!\uparrow}}}
& \textbf{87.0}\,{\scriptsize\textcolor{green!45!black}{\ensuremath{7.7\!\uparrow}}}
& \textbf{60.3}\,{\scriptsize\textcolor{green!45!black}{\ensuremath{8.4\!\uparrow}}}
& \textbf{40.9}\,{\scriptsize\textcolor{green!45!black}{\ensuremath{4.0\!\uparrow}}}
& \textbf{66.4}\,{\scriptsize\textcolor{green!45!black}{\ensuremath{5.3\!\uparrow}}}
& \textbf{58.0}\,{\scriptsize\textcolor{green!45!black}{\ensuremath{8.7\!\uparrow}}} \\

\midrule

$\pi_{0.5}$ / Base.
& 38.1
& 63.6
& 76.1
& \textbf{93.4}
& 86.3
& 66.1
& 78.8
& 70.0 \\

\rowcolor{green!4}
$\pi_{0.5}$ / Ours$^{\dagger}$
& 39.3\,{\scriptsize\textcolor{green!45!black}{\ensuremath{1.2\!\uparrow}}}
& 77.7\,{\scriptsize\textcolor{green!45!black}{\ensuremath{14.1\!\uparrow}}}
& \textbf{76.6}\,{\scriptsize\textcolor{green!45!black}{\ensuremath{0.5\!\uparrow}}}
& 93.1\,{\scriptsize\textcolor{black!70!black}{\ensuremath{0.3\!\downarrow}}}
& 87.1\,{\scriptsize\textcolor{green!45!black}{\ensuremath{0.8\!\uparrow}}}
& 66.7\,{\scriptsize\textcolor{green!45!black}{\ensuremath{0.6\!\uparrow}}}
& 80.1\,{\scriptsize\textcolor{green!45!black}{\ensuremath{1.3\!\uparrow}}}
& 72.9\,{\scriptsize\textcolor{green!45!black}{\ensuremath{2.9\!\uparrow}}} \\

\rowcolor{green!8}
$\pi_{0.5}$ / \textbf{Ours}
& \textbf{41.2}\,{\scriptsize\textcolor{green!45!black}{\ensuremath{3.1\!\uparrow}}}
& \textbf{81.4}\,{\scriptsize\textcolor{green!45!black}{\ensuremath{17.8\!\uparrow}}}
& \textbf{76.6}\,{\scriptsize\textcolor{green!45!black}{\ensuremath{0.5\!\uparrow}}}
& 93.2\,{\scriptsize\textcolor{black!70!black}{\ensuremath{0.2\!\downarrow}}}
& \textbf{87.5}\,{\scriptsize\textcolor{green!45!black}{\ensuremath{1.2\!\uparrow}}}
& \textbf{67.3}\,{\scriptsize\textcolor{green!45!black}{\ensuremath{1.2\!\uparrow}}}
& \textbf{80.3}\,{\scriptsize\textcolor{green!45!black}{\ensuremath{1.5\!\uparrow}}}
& \textbf{73.7}\,{\scriptsize\textcolor{green!45!black}{\ensuremath{3.7\!\uparrow}}} \\

\midrule

QwenPI / Base.
& 55.7
& 61.0
& 94.5
& \textbf{98.0}
& 97.6
& 86.2
& 79.1
& 80.2 \\

\rowcolor{green!4}
QwenPI / Ours$^{\dagger}$
& 57.9\,{\scriptsize\textcolor{green!45!black}{\ensuremath{2.2\!\uparrow}}}
& 76.4\,{\scriptsize\textcolor{green!45!black}{\ensuremath{15.4\!\uparrow}}}
& \textbf{94.7}\,{\scriptsize\textcolor{green!45!black}{\ensuremath{0.2\!\uparrow}}}
& 97.8\,{\scriptsize\textcolor{black!70!black}{\ensuremath{0.2\!\downarrow}}}
& \textbf{98.6}\,{\scriptsize\textcolor{green!45!black}{\ensuremath{1.0\!\uparrow}}}
& 86.6\,{\scriptsize\textcolor{green!45!black}{\ensuremath{0.4\!\uparrow}}}
& 79.6\,{\scriptsize\textcolor{green!45!black}{\ensuremath{0.5\!\uparrow}}}
& 83.1\,{\scriptsize\textcolor{green!45!black}{\ensuremath{2.9\!\uparrow}}} \\

\rowcolor{green!8}
QwenPI / \textbf{Ours}
& \textbf{59.9}\,{\scriptsize\textcolor{green!45!black}{\ensuremath{4.2\!\uparrow}}}
& \textbf{82.4}\,{\scriptsize\textcolor{green!45!black}{\ensuremath{21.4\!\uparrow}}}
& \textbf{94.7}\,{\scriptsize\textcolor{green!45!black}{\ensuremath{0.2\!\uparrow}}}
& 97.8\,{\scriptsize\textcolor{black!70!black}{\ensuremath{0.2\!\downarrow}}}
& \textbf{98.6}\,{\scriptsize\textcolor{green!45!black}{\ensuremath{1.0\!\uparrow}}}
& \textbf{86.9}\,{\scriptsize\textcolor{green!45!black}{\ensuremath{0.7\!\uparrow}}}
& \textbf{80.1}\,{\scriptsize\textcolor{green!45!black}{\ensuremath{1.0\!\uparrow}}}
& \textbf{84.6}\,{\scriptsize\textcolor{green!45!black}{\ensuremath{4.4\!\uparrow}}} \\

\bottomrule
\end{tabular}
\par\vspace{-12pt}
\end{table}

\paragraph{Evaluation protocol.}
We use benchmark-defined success criteria and keep
evaluation conditions and execution budgets matched
within each baseline--RESETTLE comparison.
Each pair uses identical base-policy weights, which
remain frozen throughout evaluation.
We report success rates as percentages and compute
all success-rate gains as absolute differences between
the displayed values, with gains expressed in percentage points.
The real-world evaluation uses stage
completion for the long-horizon task, as specified in detail below.

\paragraph{Recovery module.}
We train separate recovery modules for LIBERO-Plus,
Meta-World, and RoboCasa Tabletop, sharing each module
across the corresponding base policies.
LIBERO-Pro Swap reuses the LIBERO-Plus recovery module.
The two real-world policies likewise share one
recovery module trained for the real-world setting.

The recovery module uses an approximately 1B-parameter
ViT-g/16 encoder.
We initialize the encoder and auxiliary dynamics
predictor from V-JEPA-2-AC~\citep{assran2025v}, using
the checkpoint with action-conditioned pretraining on
DROID~\citep{khazatsky2024droid}, and jointly optimize
them with a residual head on expert demonstrations.
The adapted encoder constructs the same-task reference
bank.
Although recovery-module weights are shared across
base policies within each setting, intervention
thresholds are calibrated separately for each base
policy using training-data inference P95, without evaluation
outcomes. The threshold-percentile ablation varies this default;
all settings require two consecutive exceedances.
Appendix~\ref{app:recovery_training} provides training
and deployment details, parameter counts, and
simulation evaluation protocols.

\subsection{Experiments on Single-Arm Manipulation}
\label{sec:exp_libero_plus_metaworld}

\paragraph{Results on LIBERO-Plus.}
We evaluate FastWAM~\citep{yuan2026fast},
$\pi_{0.5}$~\citep{intelligence2025pi_}, and
QwenPI~\citep{ye2026starvla} on
LIBERO-Plus~\citep{fei2026libero}, comprising
10,030 cases across seven perturbation dimensions:
camera, robot, language, light, background, noise,
and layout.
All three base policies have been trained on LIBERO
training data~\citep{liu2023libero} and receive no
additional policy fine-tuning for this evaluation.
Table~\ref{tab:libero_plus_robust} reports success rates
within each perturbation dimension and across all
evaluation cases. The upper portion of Table~\ref{tab:libero_plus_robust}
provides published results for additional baselines:
OpenVLA~\citep{kim2024openvla},
OpenVLA-OFT~\citep{kim2025fine},
NORA~\citep{hung2025nora},
WorldVLA~\citep{cen2025worldvla},
UniVLA~\citep{bu2025univla},
$\pi_0$~\citep{black2024pi_0},
$\pi_0$-FAST~\citep{pertsch2025fast},
RIPT-VLA~\citep{tan2025interactive},
Abot-M0~\citep{yang2026abot},
Qwen3VL-OFT~\citep{ye2026starvla},
and VLAct~\citep{yang2026beyond}.
These results are taken from the cited works;
our paired evaluations compare FastWAM, $\pi_{0.5}$,
and QwenPI with their RESETTLE-augmented counterparts.

RESETTLE improves overall success for all three policies.
FastWAM increases from 49.3\% to 58.0\%,
$\pi_{0.5}$ from 70.0\% to 73.7\%, and QwenPI
from 80.2\% to 84.6\%, yielding absolute gains of
8.7\%, 3.7\%, and 4.4\%, respectively.
The largest gains occur under robot-initialization
perturbations: 16.7\% for FastWAM, 17.8\% for
$\pi_{0.5}$, and 21.4\% for QwenPI.
Camera perturbations show improvements across
all three policies, with gains of 9.1\%, 3.1\%,
and 4.2\%, respectively.
FastWAM improves in all seven dimensions, whereas
$\pi_{0.5}$ and QwenPI improve in six, with decreases of 0.2\% each under light perturbations.
These results demonstrate that our method can greatly improve robustness across VLA and WAM policies, with benefits varying against perturbation types.

\begin{table}[!t]
\centering
\caption{Success rates (\%) across difficulty levels on
Meta-World. Parentheses indicate the number of tasks in each
difficulty group. ``Ours'' denotes RESETTLE. Arrows indicate
absolute improvements (\%) over the corresponding baseline.}
\label{tab:metaworld_difficulty_eval}
\vspace{5pt}

\setlength{\tabcolsep}{4pt}
\renewcommand{\arraystretch}{1.12}

\begin{tabular}{@{}lccccc@{}}
\toprule
Model
& Easy (28) & Medium (11) & Hard (6)
& Very Hard (5) & Average \\
\midrule

QwenPI / Base.
& 80.36
& 57.27
& 45.00
& 60.00
& 60.66 \\

\rowcolor{green!8}
QwenPI / \textbf{Ours}
& \textbf{83.57}\,{\scriptsize\textcolor{green!45!black}{\ensuremath{3.21\!\uparrow}}}
& \textbf{62.73}\,{\scriptsize\textcolor{green!45!black}{\ensuremath{5.46\!\uparrow}}}
& \textbf{51.67}\,{\scriptsize\textcolor{green!45!black}{\ensuremath{6.67\!\uparrow}}}
& \textbf{66.00}\,{\scriptsize\textcolor{green!45!black}{\ensuremath{6.00\!\uparrow}}}
& \textbf{65.99}\,{\scriptsize\textcolor{green!45!black}{\ensuremath{5.33\!\uparrow}}} \\

\midrule

SmolVLA / Base.
& 79.64
& 41.82
& 83.33
& 32.00
& 59.20 \\

\rowcolor{green!8}
SmolVLA / \textbf{Ours}
& \textbf{83.57}\,{\scriptsize\textcolor{green!45!black}{\ensuremath{3.93\!\uparrow}}}
& \textbf{46.36}\,{\scriptsize\textcolor{green!45!black}{\ensuremath{4.54\!\uparrow}}}
& \textbf{90.00}\,{\scriptsize\textcolor{green!45!black}{\ensuremath{6.67\!\uparrow}}}
& \textbf{42.00}\,{\scriptsize\textcolor{green!45!black}{\ensuremath{10.00\!\uparrow}}}
& \textbf{65.48}\,{\scriptsize\textcolor{green!45!black}{\ensuremath{6.28\!\uparrow}}} \\

\bottomrule
\end{tabular}
\par\vspace{-12pt}
\end{table}

\begin{table}[!t]
    \centering

    \begin{minipage}[t]{0.49\textwidth}
        \vspace{0pt}
        \centering

        \caption{Performance on RoboCasa Tabletop.
        ``Ours'' denotes RESETTLE.}
        \label{tab:robocasa_tabletop}
        \vspace{5pt}

        \renewcommand{\arraystretch}{1.12}

        \begin{tabular*}{0.88\linewidth}
            {@{\extracolsep{\fill}}cc@{}}
            \toprule
            Model & Success (\%) \\
            \midrule

            LDA / Base.
            & 50.33 \\

            \rowcolor{green!8}
            LDA / \textbf{Ours}
            & \textbf{52.42}\,{\scriptsize\textcolor{green!45!black}{\ensuremath{2.09\!\uparrow}}} \\

            \midrule

            QwenGR00T / Base.
            & 53.67 \\

            \rowcolor{green!8}
            QwenGR00T / \textbf{Ours}
            & \textbf{60.50}\,{\scriptsize\textcolor{green!45!black}{\ensuremath{6.83\!\uparrow}}} \\

            \bottomrule
        \end{tabular*}
    \end{minipage}
    \hfill
    \begin{minipage}[t]{0.49\textwidth}
        \vspace{0pt}
        \centering

        \caption{Performance on LIBERO-Pro Swap.
        ``Ours'' denotes RESETTLE.}
        \label{tab:libero_pro_swap}
        \vspace{5pt}

        \renewcommand{\arraystretch}{1.12}

        \begin{tabular*}{0.88\linewidth}
            {@{\extracolsep{\fill}}cc@{}}
            \toprule
            Model & Success (\%) \\
            \midrule

            $\pi_{0.5}$ / Base.
            & 15.00 \\

            \rowcolor{green!8}
            $\pi_{0.5}$ / \textbf{Ours}
            & \textbf{24.00}\,{\scriptsize\textcolor{green!45!black}{\ensuremath{9.00\!\uparrow}}} \\

            \midrule

            GPT-5.5 + $\pi_{0.5}$ / Base.
            & 42.00 \\

            \rowcolor{green!8}
            GPT-5.5 + $\pi_{0.5}$ / \textbf{Ours}
            & \textbf{50.00}\,{\scriptsize\textcolor{green!45!black}{\ensuremath{8.00\!\uparrow}}} \\

            \bottomrule
        \end{tabular*}
    \end{minipage}

\end{table}

\paragraph{Results on Meta-World.}
We evaluate QwenPI and SmolVLA~\citep{shukor2025smolvla}
on all 50 Meta-World tasks~\citep{yu2020meta},
without introducing additional perturbations.
The evaluation includes 28 easy, 11 medium, 6 hard,
and 5 very hard tasks.
Each configuration is evaluated over 10 episodes per
task, with a maximum of 400 environment steps per
episode, yielding 500 episodes in total.
In Table~\ref{tab:metaworld_difficulty_eval},
Average denotes the unweighted mean of the four
difficulty-group success rates.

RESETTLE improves performance in every difficulty
group for both policies.
The macro-average success rate increases from
60.66\% to 65.99\% for QwenPI and from
59.20\% to 65.48\% for SmolVLA, yielding
absolute gains of 5.33\% and 6.28\%, respectively.
QwenPI achieves its largest improvement on hard tasks,
increasing from 45.00\% to 51.67\%, a gain of 6.67\%.
SmolVLA achieves its largest improvement on very hard
tasks, increasing from 32.00\% to 42.00\%, a gain
of 10.00\%.
These results show that recovery also benefits
standard manipulation tasks without added perturbations.
The gains across both policies further demonstrate
that a shared recovery module can support different
base policies without policy-specific recovery training.

\subsection{Experiments on RoboCasa Tabletop}
\label{sec:exp_robocasa}
We evaluate LDA~\citep{lyu2026lda} and
QwenGR00T~\citep{ye2026starvla,bjorck2025gr00t}
on the 24-task RoboCasa Tabletop GR1
suite~\citep{nasiriany2024robocasa}.
Both policies have been trained on RoboCasa training
data and share the same RoboCasa-specific recovery module.
As shown in Table~\ref{tab:robocasa_tabletop},
RESETTLE increases LDA success from 50.33\% to
52.42\%, an absolute gain of 2.09\%.
For QwenGR00T, success increases from 53.67\% to
60.50\%, an absolute gain of 6.83\%.
These results extend the benefits of action-level
recovery to another manipulation environment and
two additional policy backbones, while retaining
the original base-policy weights.

\subsection{Hierarchical Execution on LIBERO-Pro Swap}
\label{sec:exp_harness}

We further evaluate RESETTLE within
Harness VLA~\citep{zhang2026harness}, using a
GPT-5.5 planner and a frozen $\pi_{0.5}$ policy.
The planner uses task instructions, observations,
and memory to coordinate execution and respond
to feedback, while the VLA provides local
manipulation control.
We retain the planner and its execution framework
and add RESETTLE at the low-level policy's
action-execution interface to enable local recovery.

All four configurations in
Table~\ref{tab:libero_pro_swap} are evaluated on
the same 100 LIBERO-Pro Swap trials~\citep{zhou2025libero},
comprising 10 tasks with 10 seeds per task.
For direct $\pi_{0.5}$ execution, RESETTLE increases
success from 15.00\% to 24.00\%, an absolute
gain of 9.00\%.
With the GPT-5.5 planner, success increases from
42.00\% to 50.00\%, an absolute gain of 8.00\%.
The improvement within Harness VLA shows that local
corrective control remains useful when a high-level
planner already coordinates execution.
Together with the direct-policy result, this supports
the use of RESETTLE in both standalone policy execution
and hierarchical manipulation systems.

\subsection{Real-World Manipulation Experiments}
\label{sec:exp_real_world}
As shown in Figure~\ref{fig:exp_setting}, by using a Franka Research 3 robot, we evaluate RESETTLE on four single-arm manipulation tasks: `Press-button',
`Cube-up', `Carrot-in-pot', and `Use Spoon'.
The first three are short-horizon tasks evaluated
by binary success; `Use Spoon' is a long-horizon
task evaluated by stage completion.
We compare QwenPI~\citep{ye2026starvla} and
VLAct~\citep{yang2026beyond} with their respective
RESETTLE-augmented variants, keeping the base-policy
weights frozen within each comparison.
Both policies use a Qwen3-VL
backbone~\citep{bai2025qwen3}.
Our real-world VLAct implementation uses a
GR00T action head~\citep{bjorck2025gr00t}.
The two policies share the same real-world recovery
module. For each policy, a P95 threshold is calibrated
separately using offline inference on training data, and
intervention requires two consecutive threshold exceedances.

Each of the four configurations is evaluated over 20 trials per task with randomized initial arm configurations and object layouts, yielding 80 trials per configuration and 320 trials in total.
Policy and recovery inference runs on NVIDIA
RTX 4090D GPUs.
Detailed hardware, training, and evaluation settings
are provided in
Appendix~\ref{app:real_world_experiments}.

\begin{table}[t]
\centering
\caption{\textbf{Real-world performance.}
Performance (\%) over 20 trials per task and method.
Press-button, Cube-up, and Carrot-in-pot report success rate (SR), while Use Spoon reports normalized stage completion (SC). ``Ours'' denotes RESETTLE. Avg. is the mean across four tasks.}
\label{tab:real_world_results}
\vspace{5pt}

\setlength{\tabcolsep}{3.5pt}
\renewcommand{\arraystretch}{1.12}

\begin{tabular}{@{}lccccc@{}}
\toprule
Model
& Press-button
& Cube-up
& Carrot-in-pot
& Use Spoon
& Avg. \\
& SR
& SR
& SR
& SC
& \\
\midrule

QwenPI / Base.
& 90
& 55
& 85
& 36.7
& 66.67 \\

\rowcolor{green!8}
QwenPI / \textbf{Ours}
& \textbf{100}
& \textbf{70}
& \textbf{95}
& \textbf{50}
& \textbf{78.75} \\

\midrule

VLAct / Base.
& 90
& 85
& 90
& 80
& 86.25 \\

\rowcolor{green!8}
VLAct / \textbf{Ours}
& \textbf{100}
& \textbf{90}
& \textbf{95}
& \textbf{86.7}
& \textbf{92.92} \\

\bottomrule
\end{tabular}

\end{table}

\paragraph{Results.}
Table~\ref{tab:real_world_results} shows that our method improves on all four tasks for both policies. The average task metric rises from 66.67\% to 78.75\% for QwenPI and from 86.25\% to 92.92\% for VLAct, with absolute gains of 12.08\% and 6.67\%, respectively. These gains cover binary success on the three short-horizon tasks and the long-horizon task `Use Spoon'. 
Both augmented policies succeed in all 20 Press-button trials and 19 of 20 Carrot-in-pot trials, using unchanged base-policy weights.
\section{Analysis}

\textbf{Analysis of when to intervene.}
Figure~\ref{fig:analysis_action} compares QwenPI's action
disagreement under perturbations in layout and robot initial states. Distributions are computed over policy decisions, with evaluation decisions grouped by rollout outcome.
Training-data inference and successful executions align in the
dominant low-disagreement region, whereas failures have heavier
tails beyond the training-derived P95 threshold. This supports
training-based calibration without evaluation outcomes.
With two consecutive P95 exceedances, disagreement triggers
on 77.89--100.00\% of failed trajectories versus
9.22--42.81\% of successful trajectories across the seven
LIBERO-Plus perturbation categories
(Appendix Table~\ref{tab:intervention_trigger_analysis}).
Persistent disagreement is informative but not a definitive
failure label, motivating brief correction followed by re-observation.
QwenPI's calibration distributions remain similar with
5 training trajectories, 100 trajectories, or the full
training set (Appendix Table~\ref{fig:threshold_calibration_size}),
supporting calibration from a small subset.
Single- and two-seed inference have the same measured latency
at the reported precision: approximately 297ms for FastWAM
and 109ms for QwenPI.
Thus monitoring adds negligible measured latency
in these implementations.

\textbf{Analysis of how to recover.}
Table~\ref{tab:libero_plus_robust} isolates the contribution
of the visual residual. State-servo-only recovery raises
aggregate success from 49.3\% to 55.9\% for FastWAM,
70.0\% to 72.9\% for $\pi_{0.5}$, and 80.2\% to 83.1\%
for QwenPI. Adding the guarded residual yields 58.0\%,
73.7\%, and 84.6\%, respectively, and improves or matches
servo-only recovery in every perturbation dimension at the
reported precision. Robot-initialization perturbations show
the largest additional gains: 4.6\%, 3.7\%, and 6.0\%. The servo prior supplies a reference-directed motion,
while the residual uses current and reference observations
to account for scene differences that robot-state displacement
alone cannot capture. Its guard constrains the adjustment
relative to the prior. These results support visual refinement,
without establishing collision avoidance.
Paired QwenPI rollouts with the same task, layout, and seed in
Appendix~\ref{app:ood_action_analysis} show how pose adjustments
and disengagement from local obstructions enable the base
policy to resume grasping, placement, and contact manipulation.

\begin{table}[t]
    \centering
    \caption{\textbf{Monitoring and recovery computation latency.}
    Monitor, Recovery, and Total are in milliseconds; lower is better.
    Total is Monitor + Recovery under the measured QwenPI-based
    configurations. Monitor denotes additional latency beyond
    nominal policy inference. Relative latency is Total divided by
    RESETTLE's 83.6\,ms. VoLoAgent ranges give the place and grasp
    branches as lower and upper endpoints, respectively.
    LPB$^{\dagger}$ is implemented
    in our QwenPI+V-JEPA framework.}
    \label{tab:monitor_recovery_latency}
    \vspace{6pt} 
    \begin{tabular*}{\textwidth}{
        @{\extracolsep{\fill}}lcccc@{}
    }
        \toprule
        Method & Monitor & Recovery & Total $\downarrow$
        & Relative latency $\downarrow$ \\
        \midrule
        \textbf{RESETTLE (Ours)}
        & $\approx$\textbf{0.0} & \textbf{83.6} & $\approx$\textbf{83.6}
        & $\mathbf{1.0}\times$ \\
        Agentic Robot & 107.0 & 109.0 & 216.0 & $2.6\times$ \\
        VoLoAgent & 107.0 & 215.0--1193.0 & 322.0--1300.0
        & $3.9$--$15.6\times$ \\
        LPB$^{\dagger}$ & 56.2 & 423.5 & 479.7 & $5.7\times$ \\
        \bottomrule
    \end{tabular*}
    \vspace{-12pt}
\end{table}

\paragraph{Monitoring and recovery latency.}
Table~\ref{tab:monitor_recovery_latency} compares QwenPI-based
implementations of RESETTLE, Agentic Robot~\citep{yang2025agentic},
VoLoAgent~\citep{chen2026volo}, and LPB~\citep{sun2026latent}.
Both agentic methods use Qwen3-VL-4B-Instruct monitoring;
LPB is adapted to our QwenPI+V-JEPA framework.
Total combines monitoring and recovery computation,
excluding nominal inference, polling delays, and physical execution.
RESETTLE produces one action; VoLoAgent prepares a trajectory.
Our measured monitoring overhead is approximately zero, with
83.6\,ms recovery. Total latency is 61.30\%
lower than Agentic Robot's, 82.57\% lower than adapted LPB's,
and 74.04--93.57\% lower than VoLoAgent's place and grasp branches.
Direct correction avoids online planning and latent-dynamics
optimization. VLM monitoring trades computation against timeliness:
frequent checks cost more; sparse polling delays correction,
and asynchronous checks may use stale observations.
RESETTLE checks at nominal decisions; response remains
limited by the decision rate and two-consecutive-exceedance
criterion. Total measures computation, not end-to-end response time.

\paragraph{Failure cases.}
Spatial or instruction-following errors may not trigger
intervention when the policy produces consistent action proposals.
Even after intervention, recovery can fail due to unresolved
physical configurations or mismatched goals.
Adding progress and contact signals and selectively invoking
planning could mitigate these failures, but would incur additional sensing and computation costs. Given our focus on efficient intervention and recovery,
we do not explore these extensions in this work.
Appendix~\ref{app:intervention_detection} provides trigger statistics, ablations, latency details, and failure cases.
\section{Conclusion}
We introduced RESETTLE, a model-agnostic framework for
efficient action-level recovery with frozen robot policies.
It combines disagreement-based intervention, retrieval
of successful same-task references, and corrective control
through a state-servo prior with a guarded visual residual.
Each intervention executes one corrective action before
returning the control to the base policy, requiring neither
base-policy fine-tuning nor online trajectory optimization
or additional VLM reasoning calls for recovery.
Experiments across six policies and three manipulation
benchmarks demonstrate that our method can greatly improve task completion. Moreover, the integration of our method with Harness VLA shows that the local recovery complements high-level planning. Our analyses support action disagreement as an informative and low-overhead monitoring signal, and demonstrate the benefit of adding a guarded visual residual to reference-derived state-servo control. Together, these findings highlight that the reference-guided
corrections can serve as a practical way to improve robotic execution while preserving the capabilities of the deployed policy.

\bibliography{iclr2027_conference}
\bibliographystyle{iclr2027_conference}

\appendix
\clearpage
\section{Qualitative Analysis of Base-Policy Failures and Recovery Behaviors}
\label{app:ood_action_analysis}

\begin{figure}[!htbp]
    \centering
    \includegraphics[width=\linewidth,height=0.82\textheight,keepaspectratio]{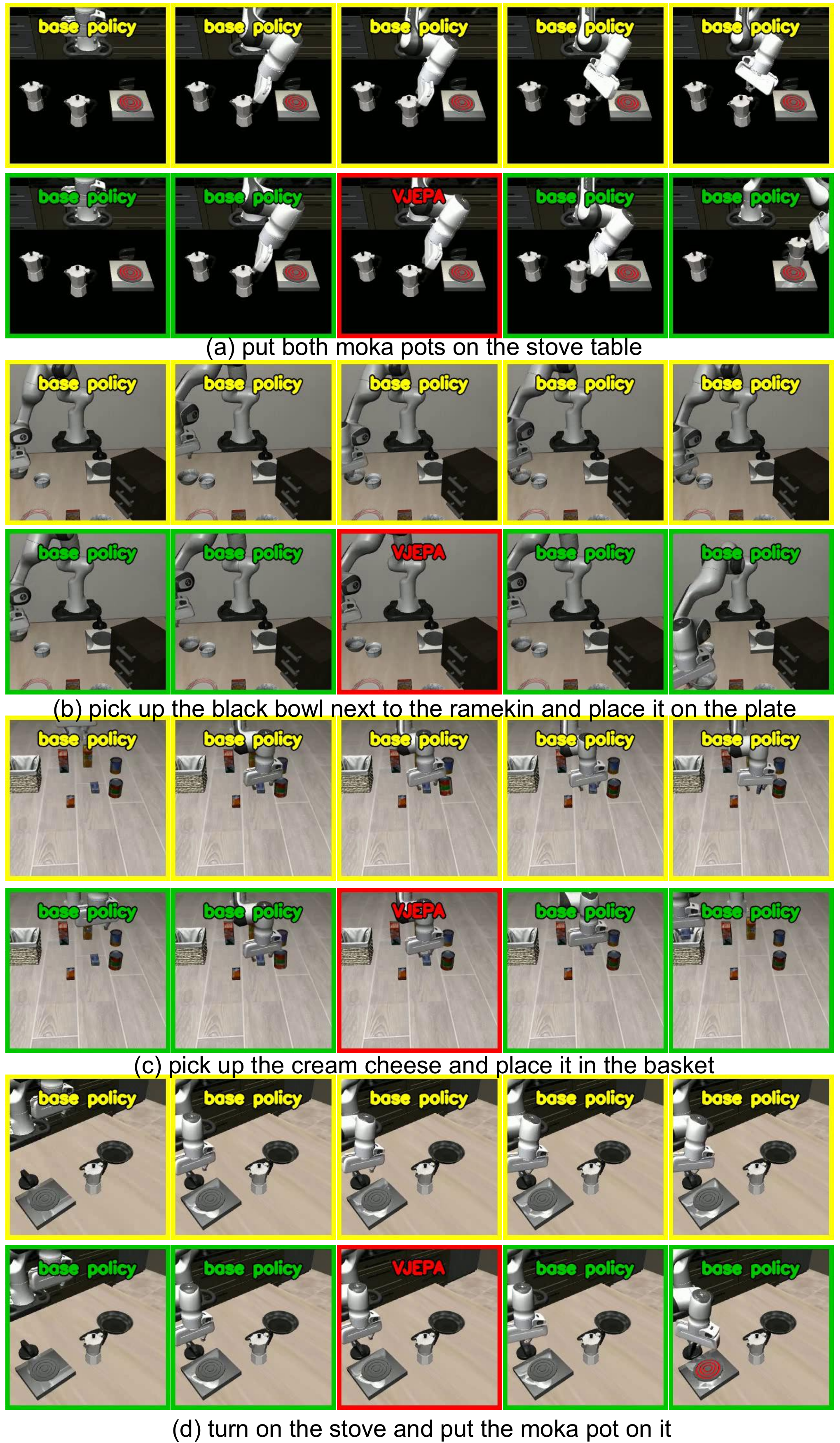}
    \caption{\textbf{Paired grasp and contact recovery cases.}
    QwenPI baseline and QwenPI+RESETTLE use the same task,
    layout, and seed in each pair. Panels (a)--(d) show
    grasp-pose correction, regrasping after a drop, recovery
    from an obstructed grasp, and correction during stove-knob
    operation. The upper yellow sequence is the baseline;
    the lower sequence alternates green base-policy execution
    with red RESETTLE interventions. Red frames labeled
    ``VJEPA'' denote the recovery controller using V-JEPA
    representations, reference retrieval, and guarded residual
    control. Frames progress from left to right.}
    \label{fig:recovery_cases_grasp_contact}
\end{figure}

\begin{figure}[!htbp]
    \centering
    \includegraphics[width=\linewidth,height=0.82\textheight,keepaspectratio]{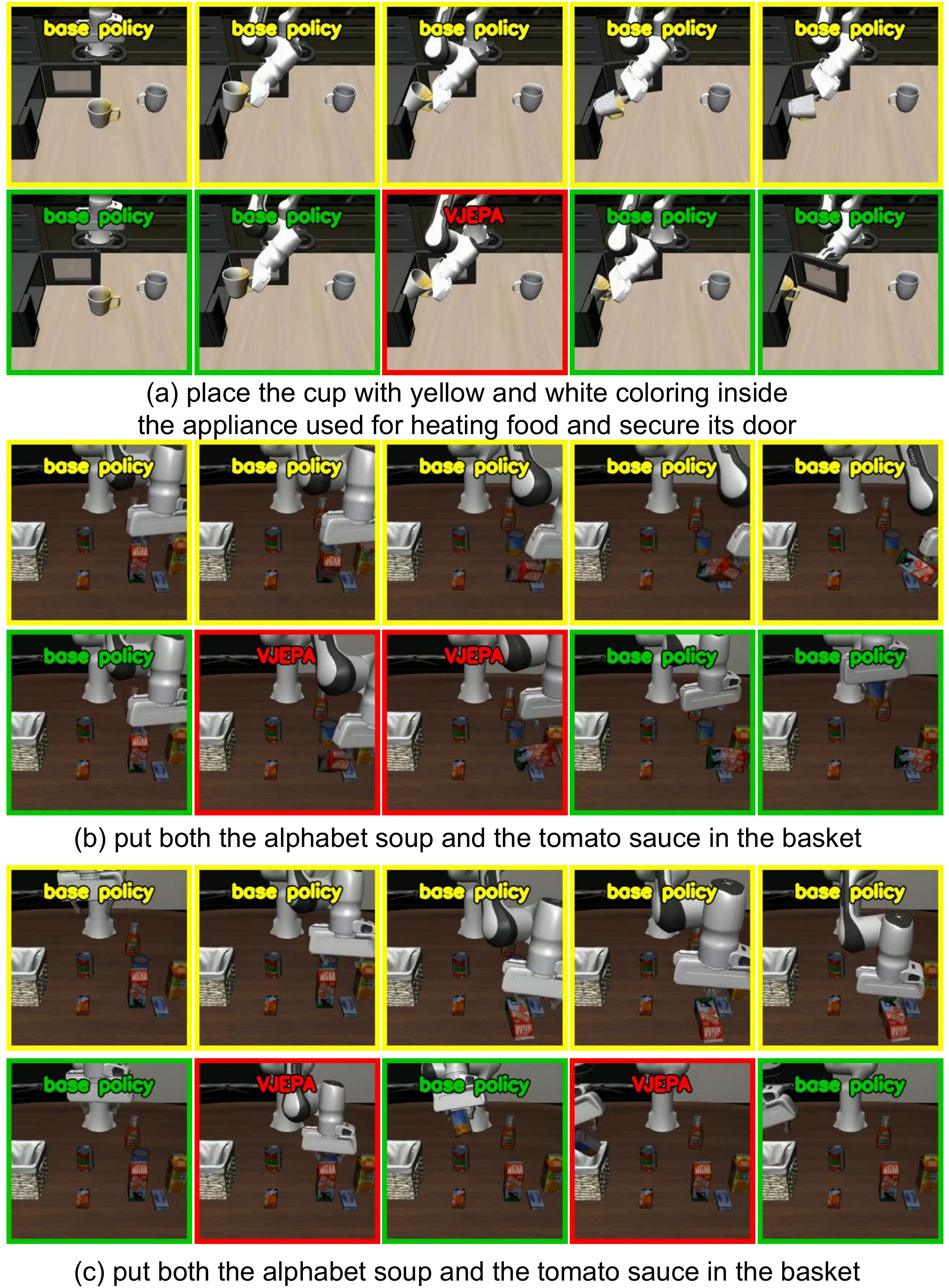}
    \caption{\textbf{Paired placement and grasp correction cases.}
    Panels (a)--(c) show recovery during obstructed cup placement,
    correction after unintended object grasping, and grasp-pose
    adjustment that avoids an unfavorable dropped-object state.
    Each pair matches the task, layout, and seed, with yellow
    baseline frames above and green base-policy/red recovery
    frames below, following
    Figure~\ref{fig:recovery_cases_grasp_contact}.
    Panel (c) includes recovery at two stages, interleaved with
    base-policy execution. Selected frames show local corrections
    and subsequent progress rather than every control step.}
    \label{fig:recovery_cases_placement_regrasp}
\end{figure}

We compare QwenPI with and without RESETTLE on LIBERO-Plus,
matching the task, layout, and evaluation seed within each pair.
Figures~\ref{fig:recovery_cases_grasp_contact}
and~\ref{fig:recovery_cases_placement_regrasp} show seven examples
of base-policy errors and subsequent correction. Frames are
selected in temporal order within each rollout; paired columns
do not imply synchronized execution after intervention.
These examples illustrate how execution errors can leave the
policy in unfamiliar configurations where its actions fail to
advance the task. They motivate recovery from OOD execution
states without identifying a precise distribution boundary
from images alone.

\paragraph{1. Correcting an unfavorable grasp pose.}
In Figure~\ref{fig:recovery_cases_grasp_contact}(a), the baseline
approaches a moka pot with an unsuitable gripper orientation
and continues ineffective grasp attempts. RESETTLE adjusts
the gripper pose, after which the base policy establishes a
grasp and transfers the pot toward the stove. The intervention
restores a useful configuration for subsequent policy execution.

\paragraph{2. Regrasping after object loss.}
In Figure~\ref{fig:recovery_cases_grasp_contact}(b), the bowl
drops during grasping, and the baseline does not effectively
reacquire it. A corrective pose adjustment allows the base
policy to reapproach and grasp the bowl, then continue toward
the plate. This case shows that a drop need not preclude
recovery when the resulting object configuration still permits
further manipulation.

\paragraph{3. Disengaging from an obstructed grasp.}
In Figure~\ref{fig:recovery_cases_grasp_contact}(c), nearby
objects obstruct the cream-cheese grasp, and the baseline
continues moving locally without completing acquisition.
RESETTLE changes the gripper configuration to relieve the
obstruction; the resumed policy then lifts the target and
transports it to the basket.

\paragraph{4. Correcting contact during knob operation.}
In Figure~\ref{fig:recovery_cases_grasp_contact}(d), the baseline
remains at an ineffective contact pose around the stove knob.
Recovery adjusts the interaction pose, enabling the base
policy to continue the rotation. The illuminated burner in
the final frame indicates successful activation of the stove.

\paragraph{5. Resolving an obstructed placement.}
In Figure~\ref{fig:recovery_cases_placement_regrasp}(a), the cup
becomes obstructed at the microwave opening. The baseline
continues from this unfavorable pose, leaving the cup on the
table instead of inside the microwave. RESETTLE adjusts the
placement configuration, allowing the base policy to insert
the cup and proceed to closing the door.

\paragraph{6. Correcting an unintended grasp interaction.}
In Figure~\ref{fig:recovery_cases_placement_regrasp}(b), the
baseline engages an unintended object, tips it, and continues
into an awkward arm--object configuration. Corrective motion
and gripper realignment allow the resumed policy to grasp the
intended object. This illustrates local recovery from a grasp
error; it does not establish general correction of semantic
target-selection errors.

\paragraph{7. Preserving a usable grasp configuration.}
In Figure~\ref{fig:recovery_cases_placement_regrasp}(c), an
unsuitable grasp causes the baseline to lose the object into
a configuration that frustrates further acquisition.
RESETTLE adjusts the gripper pose, after which the policy lifts
the target. A later intervention occurs near the basket before
policy execution resumes. The sequence illustrates repeated
local correction during grasping and transfer, with control
returned to the base policy between interventions.

Across these cases, recovery supports pose realignment,
disengagement from local obstruction, and renewed grasping
or placement. Each intervention executes a corrective control
action; the base policy performs the subsequent task motions
from a fresh observation. The figures show the combined
controller's behavior and do not isolate the contributions of
retrieval, the servo prior, and the visual residual. Together
with the failures in Appendix~\ref{app:failure_cases}, they
indicate that recovery depends on the particular physical
configuration and intervention timing: contact or object loss
alone does not determine whether local correction will succeed.

\section{Additional Method Details}
\label{app:resettle_method}

This section specifies the intervention criterion,
reference retrieval, recovery controller, and training
objectives of RESETTLE.
Let $o_t$, $s_t$, and $q$ denote the current observation,
measured robot state, and semantic task identifier.
The index $t$ denotes a nominal policy decision.
Each policy proposal contains $H$ actions, whereas
each recovery intervention executes one native control
tick. We use $e$ to denote the embodiment, whose
adapter specifies the state representation, action
coordinates, control scaling, and command limits.

\subsection{Disagreement-Based Intervention}
\label{sec:retrovla_ood}

\paragraph{Action disagreement.}
At each nominal decision, we query the frozen base
policy using independent random seeds under identical
observations and task conditioning:
\begin{equation}
    A_t^{(i)}
    =
    [a_{t,0}^{(i)},\ldots,a_{t,H-1}^{(i)}],
    \qquad i\in\{1,2\}.
\end{equation}
Let $\Delta_e(a,a';s)$ denote the action distance
defined by the embodiment adapter, and let $H_e\leq H$
denote the evaluated prefix length.
The disagreement score is
\begin{equation}
    d_t
    =
    \sum_{h=0}^{H_e-1}
    \frac{\gamma_e^h}
         {\sum_{j=0}^{H_e-1}\gamma_e^j}
    \Delta_e\!\left(
        a_{t,h}^{(1)},a_{t,h}^{(2)};s_t
    \right),
    \qquad 0<\gamma_e\leq1.
    \label{eq:proposal_divergence}
\end{equation}

For Cartesian policies, $\Delta_e$ is the Euclidean
distance between translational action components
in the adapter's unnormalized coordinates,
excluding rotation and gripper commands.

For RoboCasa/GR1, disagreement is computed from
normalized action proposals before conversion
to physical commands.
Let $\tilde a_{t,h}^{(i)}\in\mathbb{R}^{29}$ denote
the corresponding normalized action.
The per-step distance is
\begin{equation}
    \Delta_{\mathrm{GR1}}\!\left(
        a_{t,h}^{(1)},a_{t,h}^{(2)};s_t
    \right)
    =
    \sqrt{
        \frac{1}{29}
        \sum_{j=1}^{29}
        \left(
            \tilde a_{t,h,j}^{(1)}
            -
            \tilde a_{t,h,j}^{(2)}
        \right)^2
    }.
    \label{eq:gr1_action_disagreement}
\end{equation}
The 29 coordinates comprise seven for each arm,
six for each hand, and three for the waist.
We use proposal seeds 0 and 1,
$H_{\mathrm{GR1}}=\min(12,H)$, and
$\gamma_{\mathrm{GR1}}=0.8$.
Thus, the score measures disagreement over all
29 coordinates within the execution prefix,
with greater weight assigned to earlier actions.

Calibration and deployment use the same metric
and action normalization.
Disagreement serves as a proxy for decision uncertainty,
not a direct test of physical feasibility or task failure.

\paragraph{Threshold calibration.}
For each base policy, we compute calibration scores
through offline inference on training observations
and their associated conditioning inputs.
Each input produces two independently sampled action
proposals, whose disagreement is computed using
Eq.~\ref{eq:proposal_divergence}.
Calibration matches deployment in conditioning,
proposal sampling, action normalization, disagreement
metric, evaluated prefix length, and temporal weighting.
The predicted actions are not executed to collect
calibration rollouts, and evaluation outcomes are
not used.

Let $g$ denote a calibration group and
$\mathcal{S}_g^{\mathrm{cal}}$ its set of scores.
We set
\begin{equation}
    \tau_g
    =
    \operatorname{Quantile}_{\rho}
    \left(\mathcal{S}_g^{\mathrm{cal}}\right).
    \label{eq:task_threshold}
\end{equation}
All RESETTLE experiments use training-data inference P95
($\rho=0.95$), except the dedicated threshold-percentile ablation.
For LIBERO-Plus, each base policy uses a single P95 threshold
from its training-data inference distribution, shared across
all perturbation variants, tasks, and suites.
For the other benchmarks, we use a benchmark-wide
threshold shared across tasks.
Calibration is performed separately for each base
policy. The main text uses $\tau$ for the applicable
threshold.

\paragraph{Lightweight calibration from a small trajectory subset.}
\begin{figure}[!htbp]
    \centering
    \includegraphics[width=\linewidth]{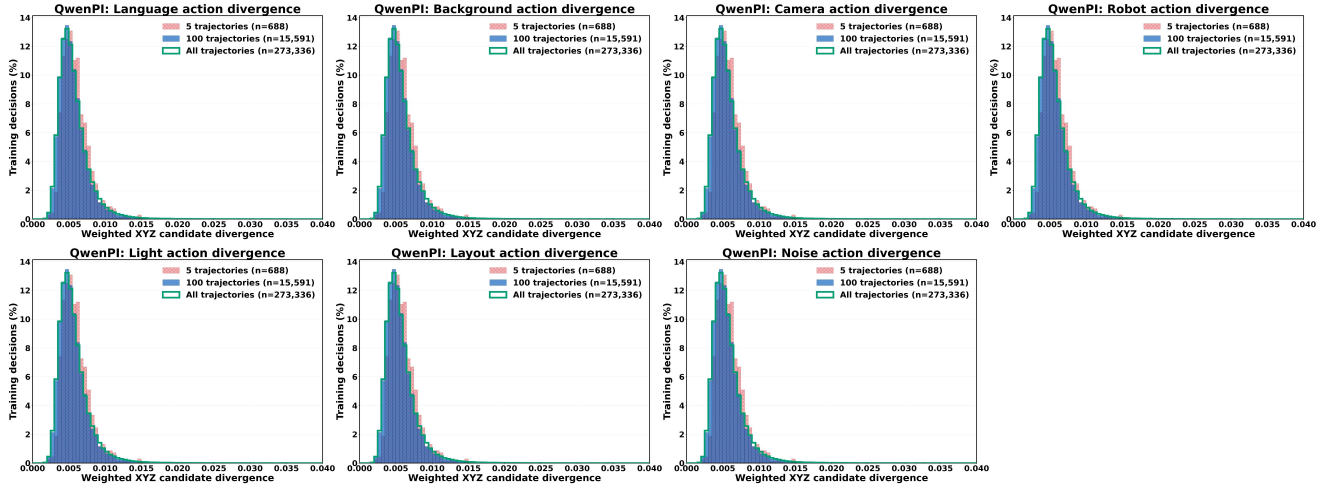}
    \caption{\textbf{Effect of calibration set size on action disagreement.}
    QwenPI distributions from 5, 100, and all training trajectories
    across seven LIBERO-Plus perturbation categories.
    The similar distribution shapes support lightweight
    threshold calibration using only five training trajectories.}
    \label{fig:threshold_calibration_size}
\end{figure}
Figure~\ref{fig:threshold_calibration_size} examines how the
amount of calibration data affects the estimated
action-disagreement distribution.
Across the seven displayed LIBERO-Plus perturbation categories,
the distribution obtained from only five training trajectories
closely follows those obtained from 100 trajectories and
the full training set, particularly in the dominant
low-disagreement region.
This observation supports a lightweight calibration procedure:
run the frozen base policy offline on observations from
five training trajectories, compute disagreement scores
using the deployment sampling protocol, and apply the
quantile rule in Eq.~\ref{eq:task_threshold}.
This requires neither additional environment interaction
nor policy optimization and substantially reduces the
number of observations processed for calibration.
The five-trajectory subset is used only for threshold
calibration; recovery-model training and reference-bank
construction retain their respective data requirements.

\paragraph{Persistent intervention criterion.}
Intervention requires two consecutive exceedances:
\begin{equation}
    r_t
    =
    \mathbf{1}[d_{t-1}>\tau_g]\,
    \mathbf{1}[d_t>\tau_g].
    \label{eq:two_consecutive_gate}
\end{equation}
This persistence requirement suppresses interventions
caused by isolated sampling fluctuations.

\paragraph{Execution protocol.}
When $r_t=0$, the first proposal follows the nominal
execution routine.
When $r_t=1$, the current proposal and any cached
action suffix are discarded, and the recovery
controller executes one native action.
The exceedance history is initialized to zero and
reset after each intervention.
Nominal execution then resumes from a fresh observation
with newly sampled proposals.
If no same-task reference is available, execution
remains with the base policy.

\subsection{V-JEPA-Based Retrieval}
\label{sec:retrovla_retrieval}

\paragraph{Reference bank.}
Let $E_\theta$ denote the adapted recovery encoder.
For single-observation encoding, the image is repeated
to form a two-frame input compatible with the encoder's
tubelet size.
The pooled representation is
\begin{equation}
    \phi_\theta(o)
    =
    \operatorname{MeanPool}\!\left(
        \operatorname{LN}
        \left(E_\theta([o,o])\right)
    \right),
\end{equation}
where $\operatorname{LN}$ normalizes token features.

For each task $q$, let $\mathcal{I}_q$ index frames
from successful training demonstrations.
After training, we construct
\begin{equation}
    \mathcal{B}_q
    =
    \{(z_i,s_i):i\in\mathcal{I}_q\},
    \qquad
    z_i
    =
    \frac{\phi_\theta(o_i)}
    {\|\phi_\theta(o_i)\|_2}.
    \label{eq:resettle_reference_bank}
\end{equation}
Each entry remains associated with its source
observation $o_i$.
The bank and online queries use the same trained
encoder and preprocessing.
Task identity restricts retrieval to demonstrations
of the same semantic task.

\paragraph{Reference selection.}
At intervention, we encode and normalize the current
observation to obtain $z_t$ and select
\begin{equation}
    i^*
    =
    \arg\min_{i\in\mathcal{I}_q}
    \left(1-z_t^\top z_i\right).
    \label{eq:nearest_success_reference}
\end{equation}
The selected observation $o_g=o_{i^*}$ supplies
visual conditioning, while its associated robot
state $s_g=s_{i^*}$ provides a reference for
corrective control.

Joint dynamics and residual-action supervision
encourages representations sensitive to task-relevant
transitions and corrective behavior.
Retrieval seeks a reference relevant to the current
execution phase without requiring explicit phase labels.
The reference is an empirical anchor of successful
behavior, not a certificate of physical reachability
or a definition of the base policy's full operating range.

\paragraph{Feasibility and demonstration references.}
Let $\mathcal{T}_{\mathrm{feasible}}(x_t)$ denote the action
sequences that can be executed from environment state $x_t$
while satisfying physical constraints and remaining compatible
with the task. Here, $x_t$ is the full environment state,
whereas $s_t$ denotes the measured robot state. A learned
policy does not explicitly characterize this feasible set:
under distribution shift it may assign high probability to
actions that cause collisions, ineffective motion, or task
failure. Demonstrated states approximate the support of
successful behavior rather than an exact feasibility boundary.
This distinction motivates using policy disagreement to detect
possible intervention needs and demonstration retrieval to
supply a correction reference.

\subsection{Reference-Guided Recovery}
\label{sec:RESETTLE_control}

\paragraph{State-servo prior.}
The embodiment-specific adapter constructs a bounded
native command
\begin{equation}
    u_t^{\mathrm{prior}}
    =
    P_e(s_t,s_g).
\end{equation}
The mapping $P_e$ accounts for the state-to-action
relationship, coordinate conventions, control interval,
scaling, and command limits.

For LIBERO, the reference displacement comprises
end-effector translation, relative rotation, and
gripper closedness:
$D_t=(p_g-p_t,R_gR_t^\top,c_g-c_t)$.
Its calibrated conversion produces the native servo
command.
Meta-World uses Cartesian translation and gripper
coordinates without rotational action dimensions.
GR1 uses the selected arm, hand, and waist action
coordinates.
The corresponding prior construction and control
constraints are embodiment-specific.

\paragraph{Visual residual.}
Let $\phi_t=\phi_\theta(o_t)$ and
$\phi_g=\phi_\theta(o_g)$.
The residual head is conditioned on the visual pair,
current robot state, previous action, its validity
indicator, and the servo prior:
\begin{equation}
\begin{split}
    (\mu_t,\log\sigma_t)
    =
    G_\eta^{(e)}\bigl(
        &\phi_t,\,
        \phi_g-\phi_t,\,
        \phi_t\odot(\phi_g-\phi_t),\\
        &s_t,\,
        a_{t-1},\,
        \nu_{t-1},\,
        b_t
    \bigr).
    \label{eq:RESETTLE_residual}
\end{split}
\end{equation}
Here, $\nu_{t-1}$ indicates whether a previous action
is available, and $b_t$ is the prior expressed in
the head's conditioning coordinates.
The output parameterizes a diagonal Gaussian
residual. Its dimensionality follows the embodiment's
action representation.

At inference, we use the predicted mean.
An embodiment-specific mask $M_e$ selects the
coordinates receiving learned corrections, and
$C_e^{\mathrm{res}}$ converts the masked prediction
to a native-action residual:
\begin{equation}
    v_t
    =
    C_e^{\mathrm{res}}(M_e\mu_t;s_t).
\end{equation}
For LIBERO, the six motion coordinates receive
learned corrections and the gripper follows the prior.
Coordinate selection and conversion for other
embodiments follow their respective action interfaces.
The residual adjusts the reference-derived prior,
not the base-policy action proposal.

\paragraph{Residual guard.}
The guard operates on corresponding groups of
native prior and residual coordinates.
For each group $b$, it checks a minimum cosine
similarity $\kappa_{e,b}^{\min}$ and a maximum
residual-to-prior norm ratio $\beta_{e,b}$.
For LIBERO, translation and rotation are checked
separately.
Other embodiments use groups defined by their
action interfaces.

A zero residual passes.
A nonzero residual paired with a negligible prior
fails, using numerical tolerance $\epsilon$.
Let $g_t$ indicate that all applicable checks pass.
The executed action is
\begin{equation}
    u_t
    =
    \begin{cases}
        \operatorname{clip}_e\!\left(
            u_t^{\mathrm{prior}}+v_t
        \right), & g_t=1,\\[2pt]
        u_t^{\mathrm{prior}}, & g_t=0.
    \end{cases}
    \label{eq:RESETTLE_guarded_control}
\end{equation}
Rejecting the residual therefore preserves the
bounded servo command.
These constraints regulate the correction relative
to the prior; they do not provide a collision-avoidance
guarantee.
Only one native action is executed before
re-observation and return to the base policy.

\subsection{Execution-Aligned JEPA Training}
\label{sec:retrovla_training}

We initialize the context encoder $E_\theta$ and
action-conditioned predictor $F_\psi$ from
V-JEPA-2-AC and jointly optimize them with the
embodiment-specific residual head $G_\eta^{(e)}$.
The target encoder $E_{\bar\theta}$ is initialized
from the context encoder and updated by exponential
moving average.
The base policy remains frozen.

\paragraph{Training samples and residual targets.}
Each training clip contains eight observations and
seven associated transitions.
A current frame supplies the observation and state,
while a future frame supplies the recovery reference.
For LIBERO, these are frames $0$ and $4$.

Action supervision is constructed according to
the data and control conventions of each embodiment.
LIBERO uses physical transition labels reconstructed
from observed state changes.
For other embodiments, the recorded action and state
representations are converted by their corresponding
adapters.
These quantities must be expressed at the same
control interval and in the same coordinates before
forming a residual target.

Let $\tilde a_0$ denote the supervised action
expressed in the residual head's output coordinates,
and let $\tilde b_0$ denote the corresponding
reference-derived prior.
The target is
\begin{equation}
    r^*
    =
    M_e\left(\tilde a_0-\tilde b_0\right).
    \label{eq:RESETTLE_residual_target}
\end{equation}
For LIBERO, the translation and rotation targets
describe a fractional motion over one native tick,
and the gripper residual target is zero.
The residual loss $\mathcal{L}_{\mathrm{res}}$
is the diagonal Gaussian negative log-likelihood.

The head-space prior and the calibrated native
servo command have distinct roles.
Physical-to-native conversion is not assumed to
commute with subtraction or clipping.

\paragraph{Execution alignment and regularization.}
The native-action loss
$\mathcal{L}_{\mathrm{native}}$ applies a Huber loss
between the candidate recovery command and the
supervised native command on the controlled
coordinates.
The candidate uses the deployed conversion and
clipping but is evaluated before hard guard
selection, allowing rejected residuals to receive
training supervision.

The penalty $\mathcal{L}_{\mathrm{gate}}$
penalizes violations of the directional and
relative-magnitude constraints.
A nonzero residual paired with a negligible prior
is penalized by its squared norm.
The saturation penalty $\mathcal{L}_{\mathrm{sat}}$
penalizes violations of the adapter's action limits
before final clipping.
All terms use the corresponding embodiment's
coordinate conventions and constraint parameters.

\paragraph{Latent dynamics prediction.}
We use the same single-step latent prediction objective
across all three benchmarks.
Each eight-frame clip provides seven adjacent transitions.
For each transition, the predictor takes the encoded
current observation, action, and robot state as inputs
and predicts the next visual representation:
\begin{equation}
    \mathcal{L}_{\mathrm{dyn}}
    =
    \frac{1}{TM}
    \sum_{j=0}^{T-1}
    \left\|
        F_\psi\!\left(E_\theta(o_j),a_j,s_j\right)
        -
        \operatorname{sg}\!\left(
            E_{\bar\theta}(o_{j+1})
        \right)
    \right\|_1,
    \label{eq:resettle_dynamics_loss}
\end{equation}
where $T=7$, $M$ is the number of latent elements
per transition, and $\operatorname{sg}$ denotes
stop-gradient.
Each prediction starts from an encoded observed frame;
predicted features are not recursively fed back
into the predictor.

The target encoder is initialized from the context
encoder and updated through an exponential moving
average:
\begin{equation}
    \bar\theta
    \leftarrow
    m\bar\theta+(1-m)\theta.
\end{equation}
It receives no gradients.
This objective encourages the shared representation
to capture action-dependent visual transitions
alongside residual-action supervision.

\paragraph{Overall objective and deployment.}
The complete objective is
\begin{equation}
\begin{split}
    \mathcal{L}
    ={}&
    \mathcal{L}_{\mathrm{res}}
    +\lambda_{\mathrm{native}}
        \mathcal{L}_{\mathrm{native}}
    +\lambda_{\mathrm{dyn}}
        \mathcal{L}_{\mathrm{dyn}}\\
    &+\lambda_{\mathrm{gate}}
        \mathcal{L}_{\mathrm{gate}}
    +\lambda_{\mathrm{sat}}
        \mathcal{L}_{\mathrm{sat}}.
    \label{eq:RESETTLE_training}
\end{split}
\end{equation}
After training, the adapted encoder constructs the
reference bank and supports online retrieval and
residual prediction.
The dynamics predictor and target encoder are
used only during training.

\section{Recovery Model Training and Implementation}
\label{app:recovery_training}

\subsection{Initialization and Network Architecture}

All benchmark-specific recovery models initialize
their context encoder and dynamics predictor from
the V-JEPA-2-AC checkpoint pretrained on DROID.
The context encoder is a ViT-g/16 with a
$16\times16$ spatial patch size and a temporal
tubelet size of two.
Training inputs contain eight RGB observations
at $256\times256$ resolution.

The action-conditioned predictor contains
24 transformer blocks, with embedding dimension
1024 and 16 attention heads.
It uses frame-causal masking and rotary positional
encoding. Extrinsics conditioning is disabled.
The context encoder and predictor are trainable;
the target encoder is updated only through EMA.

The residual head predicts a diagonal Gaussian
with an embodiment-specific output dimension.
The LIBERO head uses hidden dimension 1024,
conditioning dimension 128, depth three, and
zero dropout. Its predicted log standard deviation
is restricted to $[-5,2]$.

\paragraph{Parameter accounting.}
Table~\ref{tab:recovery_parameter_counts} reports parameter
counts for the LIBERO recovery checkpoint.
The approximately 1B designation refers to the ViT-g/16
context encoder, consistent with the V-JEPA 2 backbone
scale~\citep{assran2025v}. The full training configuration
contains approximately 2.336B instantiated parameters,
including a separate EMA target encoder, and approximately
1.324B parameters optimized by backpropagation.
The EMA encoder receives no gradients but is updated
from the context encoder through exponential moving averages.

\begin{table}[!htbp]
    \centering
    \caption{\textbf{Parameter counts for the LIBERO recovery module.}
    Trainable counts include only parameters optimized by
    backpropagation. The deployed subset consists of the context
    encoder and residual head; its trainable count is omitted
    because no optimization occurs during deployment.
    Counts exclude the base policy, reference-bank features,
    optimizer states, and non-parameter buffers.}
    \label{tab:recovery_parameter_counts}
    \vspace{5pt}
    \begin{tabular*}{\textwidth}{
        @{\extracolsep{\fill}}lrr@{}
    }
        \toprule
        Module & Total parameters & Trainable parameters \\
        \midrule
        Context encoder & 1,012,173,952 & 1,012,173,952 \\
        EMA target encoder & 1,012,173,952 & 0 \\
        Action-conditioned predictor & 305,220,992 & 305,213,824 \\
        Guarded residual-action head & 6,839,226 & 6,839,226 \\
        \midrule
        All instantiated modules & 2,336,408,122 & 1,324,227,002 \\
        Deployed subset & 1,019,013,178 & -- \\
        \bottomrule
    \end{tabular*}
\end{table}

The predictor retains an \texttt{extrinsics\_encoder}
with 7,168 parameters. These parameters are already
included in the predictor total and are frozen because
\texttt{use\_extrinsics=false}; the remaining 305,213,824
predictor parameters are trainable.
At deployment, recovery requires only the adapted context
encoder and the 6.839M-parameter residual head, totaling
approximately 1.019B parameters. Neither the dynamics
predictor nor the EMA target encoder is required online.
RESETTLE's lightweight recovery therefore refers to its
measured online computation cost and brief, selective
interventions, rather than a small total training model.
These exact counts apply to the LIBERO configuration;
embodiment-specific modules can differ on other benchmarks.

\subsection{Training Data and Temporal Sampling}

\paragraph{LIBERO.}
Training uses expert demonstrations from the
LIBERO-Object, LIBERO-Goal, LIBERO-Spatial, and
LIBERO-10 suites.
Episodes are split into training and validation
sets with fractions 0.9 and 0.1 using seed 239.
The split is performed at the episode level.

For each selected episode, the loader samples
a valid eight-frame clip at 4 FPS from the
external-camera stream.
Frames $0$ and $4$ provide the current and
reference observations.
The raw robot state is converted to end-effector
position, Euler orientation, and gripper closedness.
Physical transition labels are reconstructed
from successive observed states rather than
treated as simulator-native command labels.

\paragraph{RoboCasa/GR1.}
Training uses the 24 GR1 task datasets and
their ego-view observations.
The state and action representations contain
29 selected arm, hand, and waist coordinates.
The loader constructs valid consecutive
eight-frame windows.

\paragraph{Meta-World.}
The training dataset contains 2,500 episodes
and 204,806 raw observations.
Images are stored with their associated
four-dimensional states and actions.
The action representation contains Cartesian
translation and a gripper coordinate.

The loader samples at 4 FPS from an 80 FPS
source using a temporal stride of 20.
An eight-frame clip therefore spans
$1+7\times20=141$ source frames.
Only episodes satisfying this length requirement
provide temporal training samples, producing
17,060 valid clips under this sampling rule.

\paragraph{Image preprocessing.}
Training uses random resized crops with scale
$[0.8,1.0]$ and aspect ratio $[0.9,1.1]$.
Horizontal flipping, AutoAugment, motion shift,
and random erasing are disabled.
RGB inputs use ImageNet normalization with mean
$(0.485,0.456,0.406)$ and standard deviation
$(0.229,0.224,0.225)$.

\subsection{Embodiment-Specific Action Interfaces}

The training and deployment adapters distinguish
the three action representations:
\begin{itemize}
    \item \textbf{LIBERO:}
    seven coordinates comprising translation,
    rotation, and gripper closedness.
    Learned corrections act on the six motion
    coordinates, while the prior controls the gripper.

    \item \textbf{Meta-World:}
    four coordinates comprising Cartesian
    translation and gripper control.
    No rotational residual dimensions are used.

    \item \textbf{RoboCasa/GR1:}
    29 selected arm, hand, and waist coordinates.
    Prior construction, coordinate scaling,
    correction groups, and command bounds follow
    the GR1 action interface.
\end{itemize}

For LIBERO, the calibrated translation scales,
in meters per unit native command, are
$(0.010465,0.012178,0.010764)$.
The conversion additionally uses a calibrated
rotation-gain matrix.
The translation and rotation limits are
0.03 and 0.05 in their respective physical
displacement coordinates, and the maximum
native command magnitude is 1.0.
The minimum cosine similarities are 0.5 for
translation and 0.0 for rotation.
The maximum residual-to-prior norm ratio is 0.5.

The LIBERO conversion uses a native interval
ratio of five. This calibration does not change
the execution protocol: each intervention
executes only one native control tick.

\subsection{Base-Policy Action Chunks at Deployment}

The deployed action chunk settings for the base policies
are benchmark-specific:
\begin{itemize}
    \item \textbf{LIBERO-Plus:}
    QwenPI and $\pi_{0.5}$ use action chunks of eight steps.
    FastWAM generates 32 actions per policy call but executes
    only the first five before observing the scene again
    and replanning.

    \item \textbf{Meta-World:}
    SmolVLA and QwenPI both use action chunks of eight steps.

    \item \textbf{RoboCasa Tabletop:}
    QwenGR00T and LDA both use action chunks of 12 steps.

    \item \textbf{LIBERO-Pro Swap:}
    $\pi_{0.5}$ uses five-step action chunks in both direct
    execution and the Harness VLA route. The VLA server is
    configured to return five actions, and the evaluation
    client executes the full chunk when recovery is not triggered.
\end{itemize}
In particular, FastWAM's prediction horizon is 32 steps,
whereas its nominal execution horizon is five steps.
These settings describe base-policy execution. A RESETTLE
intervention discards pending policy actions, executes one
native recovery action, and returns control to the base
policy with a fresh observation.

\subsection{Simulation Model Settings and Evaluation Protocols}
\label{app:simulation_evaluation}

The following settings specify the evaluated task sets,
episode budgets, and evaluation seeds. Evaluation seeds are
distinct from the proposal-sampling seeds used to measure
action disagreement.

\paragraph{LIBERO-Plus.}
The QwenPI evaluation uses the checkpoint saved after
100,000 training steps and an action chunk length of eight.
Evaluation covers all tasks in \texttt{libero\_goal},
\texttt{libero\_spatial}, \texttt{libero\_object}, and
\texttt{libero\_10}, across seven perturbation categories:
background, camera, language, light, layout,
robot initialization, and noise.
Each perturbation case receives one rollout, with evaluation
seed 7. The default task selection includes the full task set,
yielding the 10,030 cases reported in the main experiments.

\paragraph{Meta-World.}
We evaluate QwenPI and SmolVLA on all 50 tasks in MT50, with 10 episodes
per task and a maximum of 400 environment steps per episode.
The base evaluation seed is 4042. This specifies a budget
of 500 episodes per evaluated configuration.

\paragraph{RoboCasa Tabletop / GR1.}
We evaluate QwenGR00T and LDA on 24 tabletop manipulation tasks with the
GR1 embodiment. Each task is assigned 50 episodes, each
limited to 720 environment steps, for a planned total of
1,200 episodes per evaluated configuration.

\paragraph{LIBERO-Pro Swap.}
Evaluation is restricted to the 10 swap tasks in
\texttt{libero\_10\_swap}. Each task is run with seeds
0--9, giving 10 episodes per task and a planned total of
100 episodes per method configuration. We evaluate direct
$\pi_{0.5}$ execution and the Codex/GPT-5.5 planner route,
each with and without recovery, using the same task-seed
schedule. In both routes, the $\pi_{0.5}$ VLA server uses
five-step action chunks, executed in full by the client
when no recovery intervention is triggered.

Within Harness VLA~\citep{zhang2026harness}, GPT-5.5 acts
through the Codex SDK as a high-level agentic planner.
It uses task instructions and interaction feedback to call
tools and decide how to use manipulation primitives.
The low-level action chunks are generated by $\pi_{0.5}$;
SAM3 supplies visual perception, and RESETTLE
performs recovery when the intervention criterion
is met in recovery-enabled configurations. GPT-5.5 does not
generate the robot's low-level action chunks. Direct
$\pi_{0.5}$ execution does not invoke the GPT-5.5 planner.

\subsection{Optimization and Training Budgets}

Training uses eight NVIDIA H800 GPUs with
distributed data parallelism, bfloat16 mixed
precision, and activation checkpointing.
We use AdamW with base learning rate
$10^{-4}$ and weight decay 0.04.
The learning-rate schedule starts and ends
at $10^{-5}$.
The context encoder uses a learning-rate
multiplier of 0.1.

For LIBERO, the predictor also uses a
learning-rate multiplier of 0.1.
The target-encoder EMA momentum increases
linearly from 0.996 to 0.9999.
The loss weights are
$\lambda_{\mathrm{native}}=1$,
$\lambda_{\mathrm{dyn}}=1$,
$\lambda_{\mathrm{gate}}=0.1$, and
$\lambda_{\mathrm{sat}}=0.05$.
No terminal-goal consistency loss is used.

\begin{table}[!htbp]
\centering
\caption{\textbf{Recovery-model training budgets.}
Training budgets for the benchmark-specific recovery models.
Batch sizes count clips. A virtual epoch denotes a fixed number
of optimizer updates rather than a complete dataset traversal.}
\label{tab:recovery_training_budget}
\vspace{5pt}

\setlength{\tabcolsep}{4pt}
\renewcommand{\arraystretch}{1.12}

\begin{tabular}{@{}lccc@{}}
\toprule
Setting
& LIBERO
& RoboCasa
& Meta-World \\
\midrule

State/action dim.
& 7
& 29
& 4 \\

Frames per clip
& 8
& 8
& 8 \\

Input resolution
& $256^2$
& $256^2$
& $256^2$ \\

Per-GPU batch
& 16
& 16
& 16 \\

Global batch
& 128
& 128
& 128 \\

Training budget
& 20k updates
& 150k updates
& 20 epochs \\

\bottomrule
\end{tabular}
\end{table}

LIBERO uses 1,000 updates per virtual epoch,
so its training budget corresponds to
20 virtual epochs.
GR1 uses 10,000 updates per virtual epoch,
making 150,000 updates equivalent to
15 virtual epochs.
Meta-World defines an epoch through the
distributed DataLoader.
Because incomplete batches are dropped,
an epoch does not imply that every valid
clip is consumed exactly once.

\subsection{Reference Bank Construction}

After training, we encode the reference observations
using the final adapted encoder.
Single-frame observations are repeated to form
two-frame inputs, followed by token normalization,
mean pooling, and $\ell_2$ normalization.
The resulting visual features have dimension 1408
and are stored with the aligned robot states and
source-observation associations.

Reference construction and temporal dynamics
training use different sampling units.
A valid dynamics clip must satisfy the temporal
window requirements, whereas an individual
demonstration frame can serve as a reference
without belonging to a valid training clip.
For example, the Meta-World bank contains
204,806 observations, while its dynamics loader
constructs 17,060 temporal clips.

Banks are partitioned by semantic task.
Online queries use the same encoder and
preprocessing as bank construction.
The encoder and residual head are deployed
for recovery; neither the dynamics predictor
nor the EMA target encoder is required online.

\section{Additional Analysis of Intervention and Recovery}
\label{app:intervention_detection}

\begin{figure}[!htbp]
    \centering
    \includegraphics[width=\linewidth]{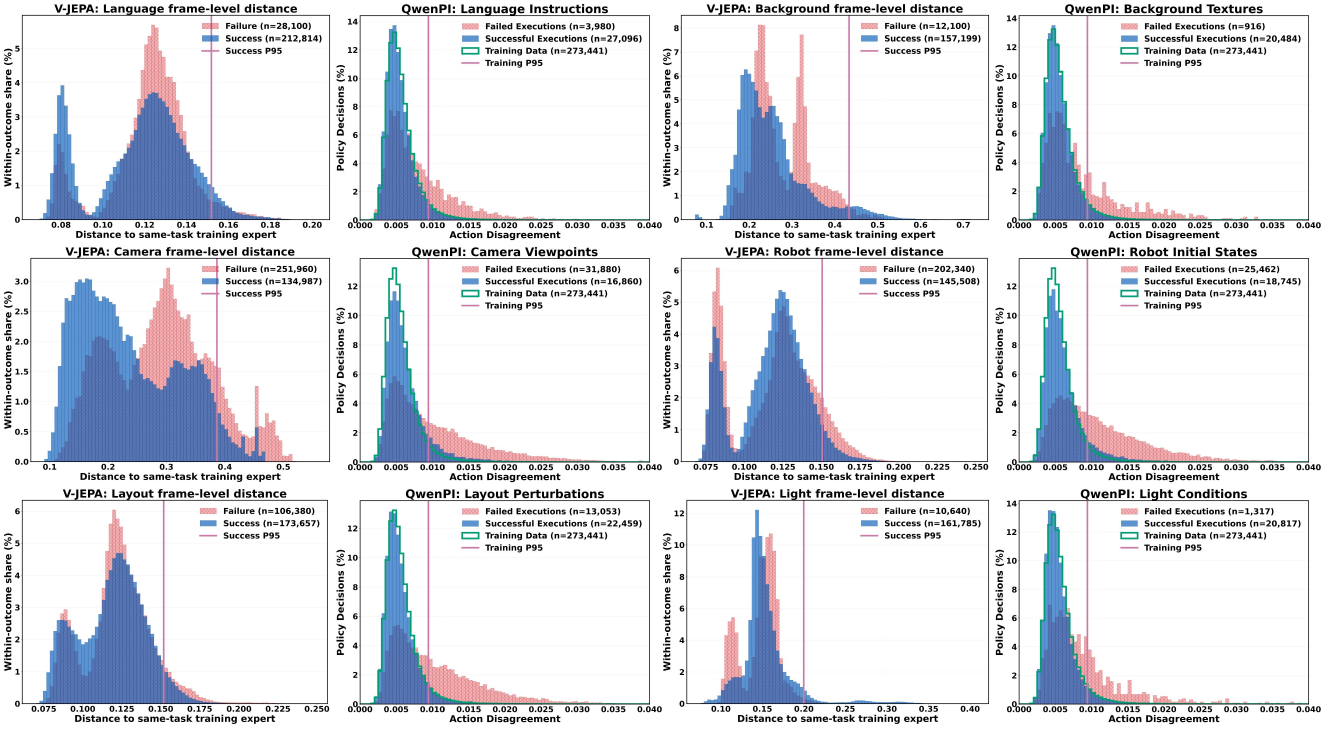}
    \caption{Feature-distance and action-disagreement distributions
    across six LIBERO-Plus perturbation categories: language instructions,
    background textures, camera viewpoints, robot initial states,
    layout perturbations, and light conditions.
    Each pair shows V-JEPA distance to same-task training references
    (left, frame-level within-outcome percentages) and QwenPI action
    disagreement (right, within-group policy-decision percentages).
    Red and blue denote failed and successful executions; green in
    disagreement panels denotes offline training-data inference.
    $n$ counts frames or policy decisions in the respective panels,
    not rollouts. Vertical lines mark ``Success P95'' for feature
    distance and ``Training P95'' for action disagreement.}
    \label{fig:distance_disagreement_comparison}
\end{figure}

\paragraph{Distribution-level comparison.}
Figure~\ref{fig:distance_disagreement_comparison} extends
the action-disagreement analysis in Figure~\ref{fig:analysis_action}
to language instructions, background textures, camera viewpoints,
robot initial states, layout perturbations, and light conditions,
and compares it with feature distance.
The distance panels report frame-level distances to same-task
training references, whereas the disagreement panels report
scores at policy decisions; each distribution is normalized
within its corresponding group. The legend counts therefore
refer to frames or policy decisions, rather than independent
evaluation rollouts. ``Training Data'' in the disagreement
panels denotes offline base-policy inference on training
observations; ``Success P95'' in the distance panels is
calibrated from successful executions.
Successful demonstrations provide useful recovery anchors, but
distance to them need not fully characterize the operating range
of a generalist policy.
The feature-distance distributions overlap substantially between
successful and failed executions, and failed executions are not
consistently farther from training references.
In contrast, the action-disagreement distributions from training
inference and successful evaluation executions closely align in
their dominant low-disagreement region, while failures exhibit
heavier high-disagreement tails.
The layout and light panels in the bottom row show the same
qualitative pattern.
These observations motivate separating reference retrieval from
intervention detection and using the stability of the policy's
own action proposals as an intervention cue.

\paragraph{Trajectory-level trigger statistics.}
Table~\ref{tab:intervention_trigger_analysis} compares
V-JEPA feature distance and action disagreement across
seven LIBERO-Plus perturbation categories, using two
consecutive P95 exceedances as the trigger criterion.
The feature-distance threshold is based on successful
executions, while the action-disagreement threshold is
calibrated from offline inference on training data,
as in Figure~\ref{fig:distance_disagreement_comparison}.
Rate denotes the percentage of trajectories with at least
one trigger, computed separately for failed and successful
executions. Count denotes the mean number of triggers
among triggered trajectories in the corresponding group.

\begin{table}[!htbp]
    \centering
    \caption{
    Failure coverage under two consecutive P95 exceedances.
    Rate: trajectories triggered (\%).
    Count: mean triggers per triggered trajectory.
    }
    \label{tab:intervention_trigger_analysis}
    \vspace{5pt}
    
    \setlength{\tabcolsep}{3pt}
    \renewcommand{\arraystretch}{1.15}

    \begin{tabular*}{\textwidth}{
        @{\extracolsep{\fill}}
        lcccc@{\hspace{14pt}}cccc@{}
    }
        \toprule
        \multirow{3}{*}{Perturbation}
        & \multicolumn{4}{c}{V-JEPA Feature Distance}
        & \multicolumn{4}{c}{\textbf{Action Disagreement (Ours)}} \\
        \cmidrule(lr){2-5}
        \cmidrule(lr){6-9}

        & \multicolumn{2}{c}{Failed}
        & \multicolumn{2}{c}{Successful}
        & \multicolumn{2}{c}{Failed}
        & \multicolumn{2}{c}{Successful} \\
        \cmidrule(lr){2-3}
        \cmidrule(lr){4-5}
        \cmidrule(lr){6-7}
        \cmidrule(lr){8-9}

        & Rate (\%) & Count
        & Rate (\%) & Count
        & Rate (\%) & Count
        & Rate (\%) & Count \\
        \midrule

        Camera
        & 26.52 & 179.3 & 10.00 & 66.9
        & \textbf{93.10} & 16.6 & 29.47 & 2.4 \\

        Robot
        & 27.98 & 104.6 & 17.44 & 28.2
        & \textbf{95.40} & 16.9 & 25.40 & 2.6 \\

        Language
        & 15.48 & 70.5 & 17.89 & 24.8
        & \textbf{77.89} & 9.6 & 9.22 & 1.9 \\

        Light
        & 21.74 & 10.2 & 9.74 & 59.4
        & \textbf{91.67} & 12.6 & 12.34 & 1.8 \\

        Background
        & 19.23 & 27.6 & 8.38 & 81.2
        & \textbf{100.00} & 11.4 & 9.82 & 1.3 \\

        Noise
        & 29.41 & 331.0 & 6.16 & 121.7
        & \textbf{99.05} & 27.4 & 42.81 & 4.9 \\

        Layout
        & 21.32 & 98.6 & 15.92 & 31.0
        & \textbf{91.64} & 15.8 & 12.77 & 2.2 \\

        \bottomrule
    \end{tabular*}
\end{table}

\paragraph{Separation between failed and successful executions.}
For feature distance, the failure-minus-success trigger-rate
gap ranges from $-2.41\%$ to $23.25\%$,
indicating limited separation under the evaluated threshold.
Action disagreement triggers on 77.89--100.00\% of failed
trajectories and 9.22--42.81\% of successful trajectories.
Its failure-minus-success trigger-rate gap is
56.24--90.18\%, showing stronger separation
in every perturbation category.

\paragraph{Trigger frequency and interpretation.}
Among trajectories triggered by action disagreement,
failed executions also produce more triggers in every
category: 9.6--27.4 versus 1.3--4.9 on average.
These patterns support persistent disagreement as a
signal of recurring execution difficulty.
However, the successful-trajectory trigger rate reaches
42.81\% under noise, showing that instability can occur
even when execution ultimately succeeds.
Disagreement therefore serves as an intervention cue
rather than a definitive failure label, motivating
RESETTLE's brief corrective interventions followed by
re-observation and a return to the base policy.

\paragraph{Additional policies and benchmarks.}
Table~\ref{tab:additional_policy_trigger_analysis} extends
the trajectory-level analysis to FastWAM on LIBERO-Plus,
LDA on RoboCasa Tabletop, and SmolVLA on Meta-World.
For each policy, the P95 threshold is calibrated from
offline inference on training data, without using
evaluation outcomes.
FastWAM uses a single threshold shared across all LIBERO-Plus
perturbation categories, tasks, and suites; LDA and SmolVLA each
use a benchmark-wide threshold shared across all tasks.
A trigger requires two consecutive policy decisions
with disagreement strictly above the applicable threshold.
The exceedance counter resets after a trigger or a
decision that does not exceed the threshold, and is
initialized separately for each trajectory.
Rate and Count follow the definitions in
Table~\ref{tab:intervention_trigger_analysis}.
The overall rows pool trajectories across all 24
RoboCasa tasks and all 50 Meta-World tasks, respectively.

\begin{table}[!htbp]
    \centering
    \caption{Action-disagreement triggers for additional policies
    and benchmarks, using training-data inference P95 thresholds
    and two consecutive exceedances with reset after each trigger.
    Rate: trajectories triggered (\%).
    Count: mean triggers per triggered trajectory.}
    \label{tab:additional_policy_trigger_analysis}
    \vspace{5pt}

    \setlength{\tabcolsep}{3pt}
    \renewcommand{\arraystretch}{1.15}

    \begin{tabular*}{\textwidth}{
        @{\extracolsep{\fill}}lcccc@{}
    }
        \toprule
        \multirow{2}{*}{Setting}
        & \multicolumn{2}{c}{Failed}
        & \multicolumn{2}{c}{Successful} \\
        \cmidrule(lr){2-3}
        \cmidrule(lr){4-5}
        & Rate (\%) & Count & Rate (\%) & Count \\
        \midrule
        \multicolumn{5}{l}{\textbf{FastWAM / LIBERO-Plus}} \\
        Camera     & 42.01 & 2.0 & 12.99 & 1.5 \\
        Robot      & 53.36 & 2.4 & 11.19 & 1.4 \\
        Language   & 68.99 & 3.5 &  8.91 & 1.4 \\
        Light      & 64.83 & 3.0 & 10.71 & 1.2 \\
        Background & 64.09 & 2.7 &  9.14 & 1.5 \\
        Noise      & 48.17 & 2.3 & 10.68 & 1.4 \\
        Layout     & 74.75 & 3.5 &  7.95 & 1.6 \\
        \midrule
        \multicolumn{5}{l}{\textbf{LDA / RoboCasa Tabletop}} \\
        Overall    & 47.32 & 2.8 & 30.13 & 1.3 \\
        \midrule
        \multicolumn{5}{l}{\textbf{SmolVLA / Meta-World}} \\
        Overall (MT50) & 23.64 & 10.7 & 7.76 & 1.7 \\
        \bottomrule
    \end{tabular*}
\end{table}

Across all reported settings, failed trajectories have
higher trigger rates and, among triggered trajectories,
more triggers on average than successful trajectories.
For FastWAM, the trigger rate is 42.01--74.75\% on
failed trajectories versus 7.95--12.99\% on successful
ones across the seven perturbation categories.
LDA exhibits a smaller separation, with rates of
47.32\% versus 30.13\% and mean counts of 2.8 versus 1.3.
For SmolVLA, the corresponding rates are 23.64\% versus
7.76\%, while triggered failures produce 10.7 triggers
on average compared with 1.7 for triggered successes.
These results extend the evidence for persistent
disagreement as a signal of recurring execution
difficulty across policies and embodiments, although
the degree of separation varies.
The untriggered failures and nonzero trigger rates on
successful trajectories also show that disagreement
is an informative intervention cue rather than a
definitive failure label, consistent with RESETTLE's
brief corrective interventions followed by re-observation
and a return to the base policy.

\paragraph{Actual recovery usage during execution.}
Table~\ref{tab:qwenpi_actual_recovery_usage} reports actual
recovery takeovers in 10,030 QwenPI+RESETTLE rollouts on
LIBERO-Plus, pooled across the four suites.
The counts are obtained from executed controller transitions
and recovery actions. Each takeover executes exactly one
native recovery action using the state-servo prior with
the guarded residual, before returning control to QwenPI.
We report mean takeover counts over all trajectories
(including those with zero takeovers) and over intervened
trajectories, alongside the fraction of trajectories with
at least one takeover.
The recovery-action share is the total number of executed
recovery actions divided by the total number of executed
base-policy and recovery actions, excluding initialization
waiting actions. All aggregate ratios are computed from
pooled counts rather than averages of per-trajectory or
per-category ratios.

All seven perturbation categories, including Robot, use
a single shared training-data inference P95 threshold and require two
consecutive exceedances, with the exceedance history reset
after each intervention. These execution statistics therefore
use the same percentile and persistence criterion as the
preceding action-disagreement trigger analyses.

\begin{table}[!htbp]
    \centering
    \caption{Actual recovery usage for QwenPI+RESETTLE on
    LIBERO-Plus using a shared training-data inference P95 threshold
    and two consecutive exceedances.
    Each takeover executes one native recovery action.
    Mean counts include either all trajectories or only
    those with at least one takeover (Intervened).
    Recovery actions (\%) denotes the share of all executed
    actions, excluding initialization waiting actions.
    Overall statistics are computed from pooled counts.}
    \label{tab:qwenpi_actual_recovery_usage}
    \vspace{5pt}

    \setlength{\tabcolsep}{3pt}
    \renewcommand{\arraystretch}{1.15}

    \begin{tabular*}{\textwidth}{
        @{\extracolsep{\fill}}lcccc@{}
    }
        \toprule
        \multirow{2}{*}{Perturbation}
        & \multicolumn{2}{c}{Mean takeovers per trajectory}
        & \multirow{2}{*}{\shortstack{Intervened\\trajectories (\%)}}
        & \multirow{2}{*}{\shortstack{Recovery\\actions (\%)}} \\
        \cmidrule(lr){2-3}
        & All & Intervened & & \\
        \midrule
        Camera     &  3.90 &  8.80 & 44.28 &  1.62 \\
        Robot      & 25.32 & 29.73 & 85.16 & 12.59 \\
        Language   &  0.07 &  2.78 &  2.34 &  0.04 \\
        Light      &  0.07 &  2.58 &  2.71 &  0.05 \\
        Background &  0.29 &  2.70 & 10.78 &  0.19 \\
        Noise      &  0.26 &  3.27 &  8.00 &  0.14 \\
        Layout     &  2.06 &  7.85 & 26.23 &  1.13 \\
        \midrule
        \textbf{Overall}
        & \textbf{4.94} & \textbf{18.08}
        & \textbf{27.31} & \textbf{2.67} \\
        \bottomrule
    \end{tabular*}
\end{table}

Recovery accounts for only 49,528 of the 1,855,166
executed actions (2.67\%), leaving 97.33\% of actions
under base-policy control; 72.69\% of trajectories
complete their execution without any takeover.
This sparse use of single-action corrections supports
RESETTLE as a lightweight recovery mechanism that
intervenes selectively when the disagreement criterion
indicates instability, while preserving base-policy
control for most of execution.
These measurements quantify the extent of recovery
control; they do not directly measure wall-clock overhead
or establish that every individual intervention is necessary.

The distribution of takeovers also varies substantially
across perturbations. Camera and Layout involve takeovers
in 44.28\% and 26.23\% of trajectories, respectively,
but recovery accounts for only 1.62\% and 1.13\% of
their actions. Thus, intervention across many trajectories
can coexist with a small share of overall control.
Robot accounts for 79.24\% of all recovery actions and
has a recovery-action share of 12.59\%; the other six
categories together have a share of only 0.67\%.
This concentration occurs under the same training-data inference
P95 threshold and two-consecutive-exceedance criterion shared
across all categories. It is consistent with robot-initialization shifts
requiring more repeated state corrections, although the
aggregate counts alone do not establish the cause.
Language, Light, Background, and Noise have recovery-action
shares of just 0.04--0.19\% under their deployed settings,
indicating that these executions rely predominantly on
the base policy. Together, these results show that
recovery usage is concentrated in a subset of executions
and perturbations, with brief corrective actions
supplementing the nominal policy.

\paragraph{Sensitivity to the calibration percentile.}
Table~\ref{tab:qwenpi_threshold_ablation} compares four
calibration percentiles for QwenPI+RESETTLE on LIBERO-Plus
with the frozen QwenPI baseline without recovery. P95 is the
default; P85, P90, and P99 are evaluated only in this ablation.
We compute P85, P90, P95, and P99 thresholds from
action-disagreement scores obtained through offline
inference on training data. For each percentile, a single
threshold is shared across all perturbation categories, tasks,
and suites.
All four settings require two consecutive exceedances
and reset the exceedance history after each intervention.
Thus, this ablation varies the calibration percentile
while retaining the same persistence criterion.

\begin{table}[!htbp]
    \centering
    \caption{\textbf{Threshold percentile ablation on LIBERO-Plus.}
    Success rates (\%) for QwenPI+RESETTLE using training-data
    inference thresholds and two consecutive exceedances.
    Baseline denotes QwenPI without RESETTLE, as reported in
    Table~\ref{tab:libero_plus_robust}.
    Total aggregates all evaluation cases, weighting each
    perturbation category by its number of cases.
    Cam.: camera; Lang.: language; BG: background; Robot: robot initialization.
    Values are rounded to one decimal place; bold marks
    the best result in each column, including ties.}
    \label{tab:qwenpi_threshold_ablation}

    \begin{tabular*}{\textwidth}{
        @{\extracolsep{\fill}}lcccccccc@{}
    }
        \toprule
        Setting
        & Cam. & Robot & Lang. & Light
        & BG & Noise & Layout & Total \\
        \midrule
        Baseline
        & 55.7 & 61.0 & 94.5 & \textbf{98.0}
        & 97.6 & 86.2 & 79.1 & 80.2 \\
        \midrule
        P85
        & 58.7 & \textbf{84.2} & 94.7 & 97.5
        & 97.7 & 79.5 & 78.8 & 83.1 \\
        P90
        & 59.2 & 83.8 & 94.3 & \textbf{98.0}
        & 97.9 & 83.3 & 79.3 & 83.9 \\
        P95
        & \textbf{59.9} & 82.4 & 94.7 & 97.8
        & \textbf{98.6} & 86.9 & \textbf{80.1} & \textbf{84.6} \\
        P99
        & 57.3 & 76.2 & \textbf{94.9} & 97.5
        & 98.5 & \textbf{89.6} & \textbf{80.1} & 83.6 \\
        \bottomrule
    \end{tabular*}
\end{table}

All four recovery settings improve aggregate success over
the 80.2\% baseline. Success increases from 83.1\% at P85
to 83.9\% at P90 and 84.6\% at P95, before declining to
83.6\% at P99. At the displayed precision, these correspond
to gains of 2.9--4.4\% over the baseline.
P95 achieves the best total, leads on Camera and Background,
and ties with P99 on Layout. Thus, P95 provides the strongest
aggregate result among the tested settings, although no
percentile is best in every perturbation category.

Robot and Noise exhibit opposing trends as the
threshold increases from P85 to P99: Robot success
decreases from 84.2\% to 76.2\%, whereas Noise success
increases from 79.5\% to 89.6\%.
All four settings exceed the 61.0\% Robot baseline, but
P85 and P90 fall below the 86.2\% Noise baseline; only
P95 and P99 improve Noise success.
A lower threshold makes the trigger condition easier
to satisfy and may enable timely correction of
robot-initialization shifts. Under sensor noise,
a stricter threshold may instead suppress interventions
caused by observation noise that do not require recovery.
These are possible explanations of the observed
success-rate trends; the ablation alone does not
establish which individual interventions are necessary.
Language, Light, and Background vary relatively little.
On Light, P90 matches the 98.0\% baseline, while the other
settings are slightly lower. The Robot--Noise tradeoff
therefore illustrates why the strongest correction in one
category need not yield the best overall threshold.

\subsection{Visual Residual Ablation}
\label{app:residual_analysis}

Table~\ref{tab:libero_plus_robust} compares state-servo-only
recovery with the full controller on FastWAM, $\pi_{0.5}$,
and QwenPI. Adding the guarded visual residual raises
aggregate success by 2.1\%, 0.8\%, and 1.5\%,
respectively. The largest additional gains occur under
robot-initialization perturbations: 4.6\%, 3.7\%, and 6.0\%.
Camera success also improves by 1.0\%, 1.9\%, and 2.0\%.
At the reported precision, the residual improves or matches
servo-only recovery in every perturbation dimension.

The state-servo prior depends on robot-state displacement
and does not account for the surrounding scene. Motion toward
a reference state may therefore be inappropriate for the
current object configuration, potentially causing collisions
or disrupting contact. The residual jointly interprets current
and reference observations to predict a scene-conditioned
adjustment, while the guard constrains its alignment and
magnitude relative to the prior. The gains support the value
of visual refinement but do not directly establish a reduction
in collisions. The paired qualitative cases illustrate
grasp-pose adjustment and recovery from local obstruction
with the full controller, without isolating the residual's
contribution.

\subsection{Monitoring and Recovery Latency Details}
\label{app:latency_details}

\paragraph{Measurement scope and implementations.}
Table~\ref{tab:monitor_recovery_latency} reports the computation
costs of monitoring and preparing a recovery output in our
QwenPI-based implementations. All configurations use QwenPI
as the base VLA. Agentic Robot~\citep{yang2025agentic} and
VoLoAgent~\citep{chen2026volo} both use
Qwen3-VL-4B-Instruct~\citep{bai2025qwen3} for monitoring.
LPB~\citep{sun2026latent} is adapted to our QwenPI+V-JEPA
framework: its monitor encodes the current image and searches
same-task expert latents, while recovery performs four
future-latent prediction steps, backpropagation, and one
action update.

Monitor measures additional computation beyond nominal policy
inference. Total sums one monitor evaluation and one recovery
computation, excluding nominal policy inference, polling delays,
and physical action execution. It does not measure full-task
duration or asynchronous response time. The outputs differ:
RESETTLE produces one corrective action, whereas VoLoAgent
prepares a trajectory.

\paragraph{Shared computation for disagreement monitoring.}
The two proposals use identical conditioning, allowing condition
encodings, latent caches, and VLM prefill computation to be
reused across seeds; only diffusion or flow-matching sampling
requires additional computation. Single- and two-seed inference
both take approximately 297\,ms for FastWAM and 109\,ms for
QwenPI in our implementation. Approximately zero additional
monitoring latency denotes no measurable increase at the
reported precision, rather than an absence of extra computation.

\paragraph{Recovery computation breakdown.}
RESETTLE's 83.6\,ms recovery comprises V-JEPA encoding
(50.8\,ms), nearest-neighbor search (32.0\,ms), state-servo
control (0.2\,ms), and the visual residual (0.6\,ms).
Encoding and retrieval dominate; generating the corrective
command takes only 0.8\,ms, including the visual residual.
Agentic Robot requires 107\,ms for verification and 109\,ms
for a QwenPI recovery proposal.
VoLoAgent's grasp pipeline takes 1193\,ms for
SAM3~\citep{carion2026sam} detection and segmentation, depth
and point-cloud processing, GraspGen~\citep{murali2025graspgen}
pose prediction, and LIBERO Cartesian trajectory planning.
Its place pipeline takes approximately 215\,ms, dominated
by SAM3 target localization and depth back-projection
(214\,ms). Cartesian trajectory generation takes less than
1\,ms and produces 11 OSC\_POSE actions; the logged
0.000\,s reflects rounding to three decimal places.
LPB's online prediction and gradient-based action update
require 423.46\,ms beyond its 56.20\,ms monitor.

\paragraph{Monitoring frequency and responsiveness.}
Frequent VLM checks increase computation, whereas polling
every several execution steps reduces invocation frequency
but allows errors to persist until the next check.
Asynchronous monitoring can reduce blocking, but observations
and decisions can still become stale while the robot moves.
RESETTLE evaluates disagreement at each nominal policy decision
and executes a brief correction before re-observation.
Its response remains subject to the policy decision rate
and two-consecutive-exceedance criterion; low additional
monitoring latency does not imply instantaneous detection.
These measurements support efficient local correction alongside
the semantic planning capabilities of high-level agents.

\subsection{Failure Cases and Recovery Limitations}
\label{app:failure_cases}

\begin{figure}[!htbp]
    \centering
    \includegraphics[width=\linewidth,height=0.82\textheight,keepaspectratio]{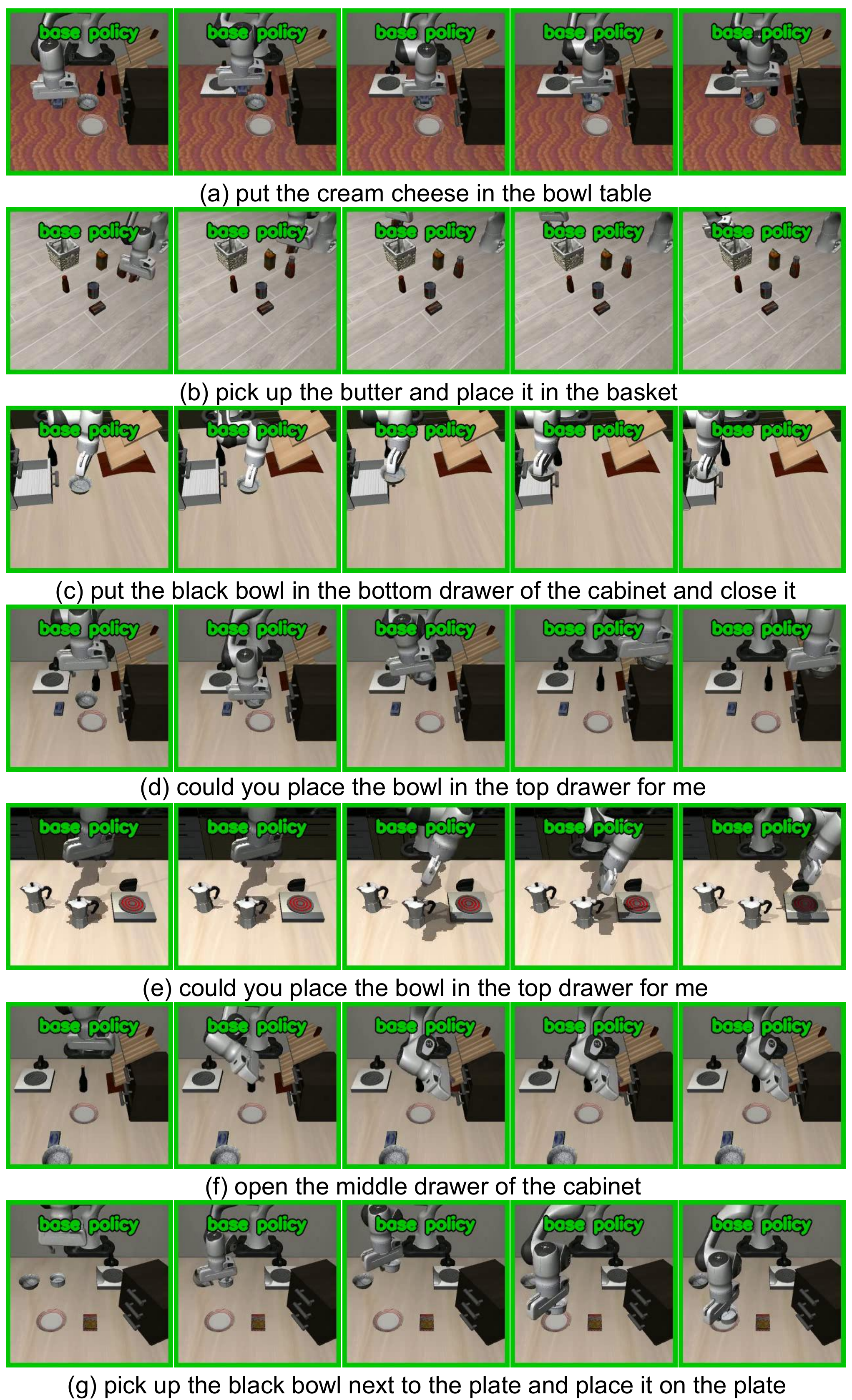}
    \caption{\textbf{Failures without a recovery trigger.}
    Seven unsuccessful rollouts with no monitoring trigger,
    ordered from top to bottom. Frames progress left to right.
    The cases include contact-induced scene changes, misplaced
    objects, obstructed motions, and incorrect grasp or task
    decisions. Consistent policy proposals can therefore
    accompany unsuccessful execution.}
    \label{fig:failure_no_trigger}
\end{figure}

\begin{figure}[!htbp]
    \centering
    \includegraphics[width=\linewidth,height=0.82\textheight,keepaspectratio]{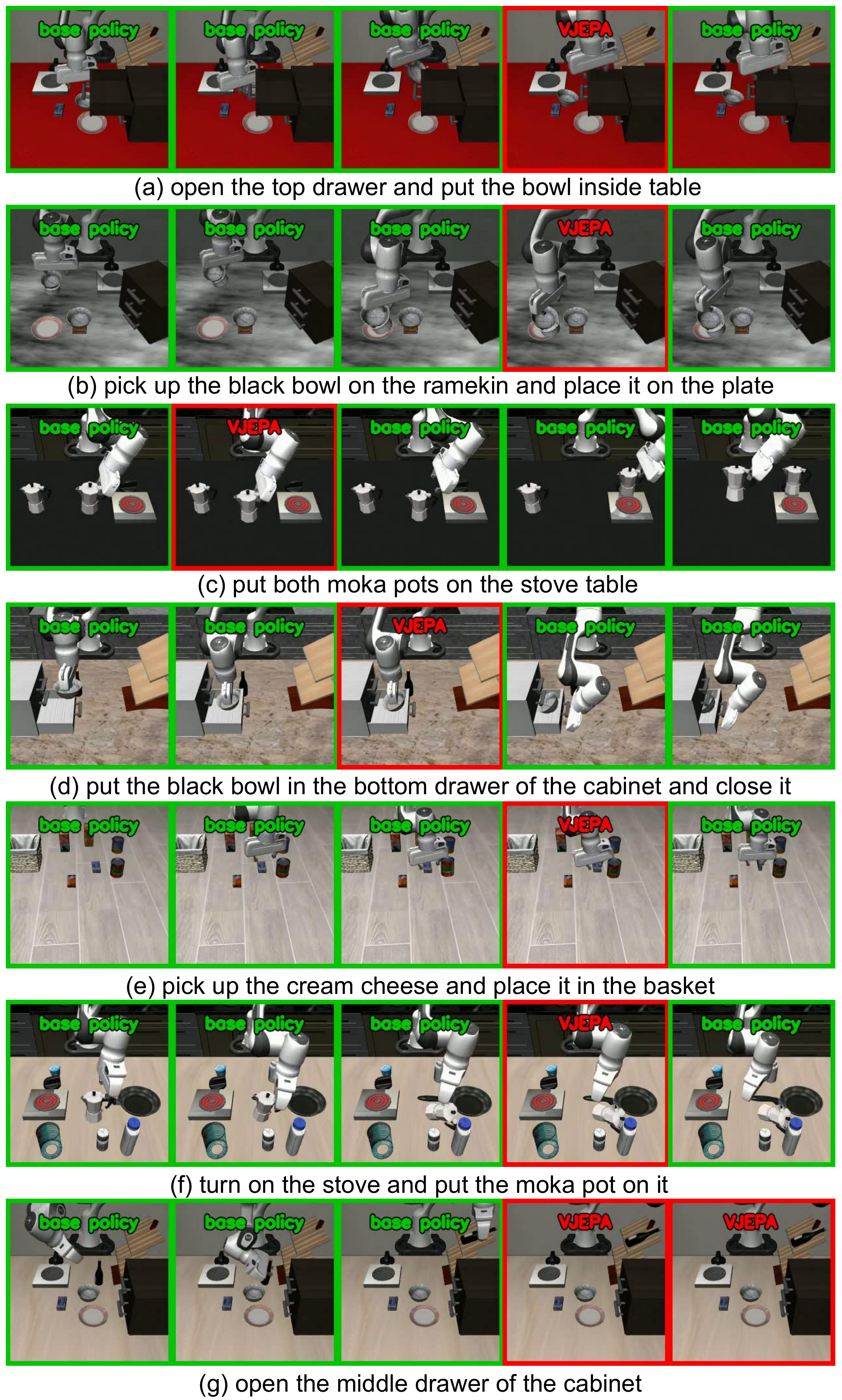}
    \caption{\textbf{Failures despite triggered recovery.}
    Seven unsuccessful rollouts in which recovery is triggered,
    ordered from top to bottom, with frames progressing left
    to right. Cases illustrate contact-induced deterioration,
    unintended grasp configurations, execution-budget exhaustion,
    and task-goal mismatch. Triggered intervention does not
    guarantee completion within the evaluation budget.}
    \label{fig:failure_after_trigger}
\end{figure}

We examine two groups of unsuccessful executions: those in
which the monitor never triggers recovery, and those in which
recovery is triggered but the task remains incomplete.
Figures~\ref{fig:failure_no_trigger}
and~\ref{fig:failure_after_trigger} each show seven cases,
numbered from top to bottom, with frames ordered left to right.
The discussion distinguishes observed behavior from possible
mechanisms. These selected cases illustrate failure modes;
they do not estimate their frequency across all evaluations.

\paragraph{Failures without intervention.}
Figure~\ref{fig:failure_no_trigger} illustrates that action
consistency does not establish physical feasibility or task
correctness. In these rollouts, disagreement does not satisfy
the intervention criterion, despite unsuccessful execution.

\begin{enumerate}
    \item \textbf{Contact-induced scene change.}
    Placing the cream into the bowl knocks the bowl out of
    position, changing the scene geometry without eliciting
    a recovery trigger.

    \item \textbf{Placement error.}
    The policy places the butter outside the basket.
    Its proposals remain consistent despite the spatial error.

    \item \textbf{Object entrapment.}
    The bowl becomes caught against the drawer, while the
    policy continues without sufficient action disagreement
    to trigger intervention.

    \item \textbf{Instruction mismatch.}
    The policy produces behavior inconsistent with the language
    instruction, but its sampled proposals remain similar.

    \item \textbf{Incorrect grasp under changed lighting.}
    The policy selects an unsuitable grasp location on the pot
    without meeting the intervention criterion.

    \item \textbf{Misaligned drawer interaction.}
    The end-effector pose is unsuitable for pulling the drawer,
    and motion becomes obstructed without a recovery trigger.

    \item \textbf{Wrong-object selection.}
    The policy grasps an unintended object while continuing
    to produce consistent action proposals.
\end{enumerate}

Cases 1, 3, and 6 highlight contact effects that need not
produce divergent proposals. Cases 2, 4, 5, and 7 show
that spatial or semantic errors can likewise remain consistent
across samples. Here, consistency describes the observed
proposal agreement, not a calibrated confidence estimate.
Disagreement alone cannot distinguish correct, stable behavior
from stable but incorrect behavior.
The successful paired cases show that some grasp and contact
errors admit local recovery.
The absence of intervention in the present cases therefore
identifies a monitoring limitation, without establishing whether
recovery would have succeeded had it been invoked.

\paragraph{Failures despite triggered recovery.}
Figure~\ref{fig:failure_after_trigger} shows failures in which
the intervention criterion is met. Triggering identifies an
opportunity for correction, but does not establish that the
current physical state is recoverable by a brief intervention.

\begin{enumerate}
    \item \textbf{Intervention after object loss.}
    The bowl collides with the drawer and falls. Recovery
    triggers after the arm has moved onward, and the earlier
    object loss is not resolved within the rollout.

    \item \textbf{Unintended multi-object grasp.}
    The gripper accidentally acquires two objects, producing
    a grasp configuration that subsequent recovery does not
    successfully correct.

    \item \textbf{Pot grasping under a finite budget.}
    Recovery is triggered following unsuccessful pot grasping,
    but the rollout reaches the maximum action count before
    task completion.

    \item \textbf{Entrapment during placement.}
    The bowl becomes caught against the drawer during placement;
    the triggered corrections do not resolve the obstruction.

    \item \textbf{Cream grasping under a finite budget.}
    Unsuccessful cream grasping is followed by intervention,
    but the remaining action budget is exhausted before the
    task is completed.

    \item \textbf{Unresolved grasp configuration.}
    An unsuitable grasp pose leaves the object in a
    configuration that recovery fails to rectify during
    the evaluated execution.

    \item \textbf{Task-goal mismatch under robot-initialization changes.}
    With a changed arm pose and visual observation, execution
    approaches a wine bottle instead of opening the requested
    drawer. Recovery does not restore instruction-consistent
    behavior in this case.
\end{enumerate}

These failures should be read alongside the successful
corrections in the paired rollouts.
Figure~\ref{fig:recovery_cases_grasp_contact}(b) shows regrasping
after a drop, whereas failure case 1 involves object loss
followed by intervention after the arm has moved onward.
Likewise, local obstructions are resolved in
Figure~\ref{fig:recovery_cases_grasp_contact}(c,d) and
Figure~\ref{fig:recovery_cases_placement_regrasp}(a), while the
drawer entrapment in failure case 4 remains unresolved.
Thus contact and object loss are not categorical limits;
the specific configuration and timing of intervention matter.
The controller can produce adjustments that enable disengagement
and regrasping, but it does not explicitly plan contact transitions
or verify recoverability. The guard bounds the residual relative
to the prior without guaranteeing collision-free motion.

Correcting the unintended grasp interaction in
Figure~\ref{fig:recovery_cases_placement_regrasp}(b) also differs
from resolving the instruction mismatch in failure case 7:
local pose correction does not provide explicit verification
of the task goal. Cases 3 and 5 additionally show that
intervention can occur without sufficient remaining budget
to complete the task. Longer execution might allow further
attempts, but these rollouts do not establish success under
an extended budget.

\paragraph{Directions for improvement and computational tradeoffs.}
On the monitoring side, task-progress, target-identity,
grasp-outcome, or contact signals could complement disagreement
and expose consistent but ineffective behavior. On the recovery
side, contact-aware reference selection or explicit planning
of release and regrasp sequences could address configurations
that repeated local corrections do not resolve. Instruction-level
verification could help identify task-goal mismatch.
These extensions may require additional sensing, supervision,
or online reasoning and optimization. The present design
focuses on selective local correction with low measured
computation cost and leaves these broader mechanisms outside
the evaluated system. A possible extension is to invoke
more expensive supervision or planning only after persistent
lack of progress or repeated unsuccessful interventions.
Whether such selective escalation improves the recovery--latency
tradeoff requires further evaluation.

\section{Real-World Experimental Details}
\label{app:real_world_experiments}

\subsection{Hardware and Observations}

Our experimental platform consists of a stationary,
table-mounted Franka Research 3 robot with seven
degrees of freedom. An Intel RealSense D435 provides
an external view of the workspace, while a
wrist-mounted Intel RealSense D405 captures close
views of the gripper and manipulated objects.
Images from both cameras are resized to
$224\times224$ for the base policy.
Figure~\ref{fig:exp_setting} shows the experimental
platform and task scenes. All real-world evaluations are run
on NVIDIA RTX 4090D GPUs.

\subsection{Tasks and Data Collection}

We evaluate three short-horizon single-arm tasks involving
button pressing, stacking, and object placement, together
with one long-horizon task involving sequential tool use:

\begin{itemize}
    \item \textbf{Press-button.}
    Press the red button.

    \item \textbf{Cube-up.}
    Pick up one cube and stack it on another cube.

    \item \textbf{Carrot-in-pot.}
    Pick up the carrot and place it inside the pot.

    \item \textbf{Use Spoon.}
    Grasp the spoon, scoop a bean from the source
    bowl, and transfer it into the target bowl.
\end{itemize}

\begin{figure}[!htbp]
    \centering
    \includegraphics[width=\linewidth]{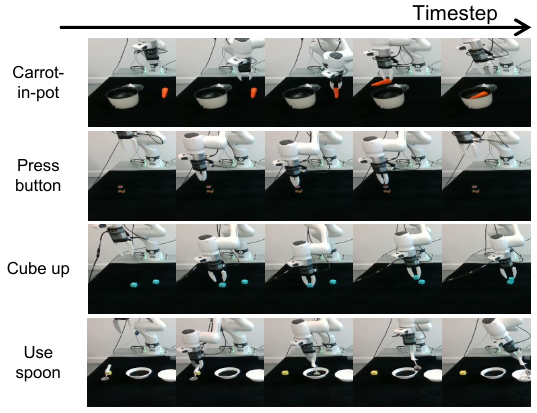}
    \caption{\textbf{Real-world training demonstrations.}
    Rows show Carrot-in-pot, Press-button, Cube-up, and Use Spoon,
    from top to bottom, with time progressing from left to right.
    Selected frames illustrate object acquisition and placement,
    button contact, stacking, and tool use. Successful training
    demonstrations support recovery-module learning and provide
    same-task references for retrieval.}
    \label{fig:real_world_training_demonstrations}
\end{figure}

Expert demonstrations are collected through human
teleoperation using a GELLO interface, with
50 demonstrations per single-arm task.
During collection, object placements and initial
robot-arm configurations are randomized across
episodes to cover different starting states.
Successful demonstrations from the training set
also provide the task-conditioned reference bank
used by RESETTLE.
Figure~\ref{fig:real_world_training_demonstrations} shows
representative sequences from the four training tasks.

\subsection{Model Training}

\paragraph{Base policies.}
We adapt QwenPI to the single-arm manipulation
setting using expert demonstrations, training for
50,000 steps on eight NVIDIA H800 GPUs.
We additionally evaluate VLAct~\citep{yang2026beyond},
which uses VLA-oriented continued pretraining to adapt
a Qwen3-VL backbone. For our real-world VLAct policy,
we pair this pretrained backbone with a GR00T action head.
For each base policy, the same frozen checkpoint is shared
between its baseline and RESETTLE-augmented variant.
Adding the recovery module does not require
further fine-tuning of the base policy.

\paragraph{Recovery module.}
We initialize the recovery encoder and auxiliary
dynamics predictor from V-JEPA-2-AC pretrained on
DROID and jointly train them with the visual
residual head on expert demonstrations.
Training combines residual action supervision,
alignment with native control commands, and
latent dynamics prediction.
The dynamics predictor is used only during training.

After training, the adapted encoder is used to
construct the reference bank from successful
training demonstrations. Intervention thresholds
are calibrated separately for QwenPI and VLAct as the P95
of action-disagreement scores obtained through offline
inference with the corresponding frozen base policy on
training data, without using evaluation outcomes. Both
policies require two consecutive exceedances, with the
exceedance history reset after each intervention.

\subsection{Evaluation Protocol}

We conduct 20 trials per task for each method,
randomizing the robot's initial arm configuration
across trials. The evaluation comprises 80 trials
per configuration and 320 trials in total across
QwenPI, QwenPI + RESETTLE, VLAct, and VLAct + RESETTLE.
Within each base-policy comparison, the baseline and
RESETTLE-augmented variant use identical base-policy
weights, observation preprocessing, task settings,
and execution budgets.

During nominal execution, the selected base policy
supplies the robot's actions. When RESETTLE triggers recovery,
pending policy actions are discarded and the
recovery controller executes one native control
action. The robot is then observed again, and
control returns to that base policy.
This intervention does not introduce a high-level
planner or additional LLM/VLM reasoning calls.

\subsection{Evaluation Metrics}

We use binary task success for the three short-horizon
tasks and a stage-completion score for the long-horizon
Use Spoon task.

\paragraph{Short-horizon tasks.}
For Press-button, Cube-up, and Carrot-in-pot, a trial
receives a score of one if the complete manipulation
goal is achieved and zero otherwise. Success requires
pressing the target button, stacking the target cube
on the other cube, or placing the carrot inside the
pot, respectively. Intermediate progress, such as
grasping the carrot without placing it in the pot,
receives no partial credit.

\paragraph{Long-horizon task.}
For Use Spoon, the three stages are acquiring
the spoon, scooping a bean, and transferring the
bean into the target bowl. Each completed stage
contributes one third of the trial score.
Completion is determined by the physical outcome:
an empty grasp or a pouring motion without a bean
does not satisfy the corresponding manipulation
criterion.

For task $q$, let $c_{q,i}\in[0,1]$ denote the
score of trial $i$. We compute
\begin{equation}
    S_q = \frac{100}{20}\sum_{i=1}^{20} c_{q,i},
    \qquad
    S_{\mathrm{avg}} =
    \frac{1}{4}\sum_{q=1}^{4} S_q.
\end{equation}
Here, $S_q$ is a success rate for short-horizon
tasks and a normalized stage-completion score
for Use Spoon. The trial score satisfies
$c_{q,i}\in\{0,1\}$ for the three short-horizon tasks
and $c_{q,i}\in\{0,1/3,2/3,1\}$ for Use Spoon.
For Use Spoon, let $K_q$ be the total
number of completed stages across the 20 trials.
Because each trial has three equally weighted stages,
$S_q=100K_q/60$.
Task metrics and their average are computed from the underlying
counts before rounding. Reported improvements are absolute
differences between the displayed values.
The reported average
gives equal weight to the four task metrics;
it is not the fraction of trials in which
all task stages are completed.

\subsection{Detailed Task-Level Results}
\label{app:real_world_results_detail}

Table~\ref{tab:real_world_results} reports improvements on
all four tasks for both policies. For QwenPI, Press-button,
Cube-up, and Carrot-in-pot success rates increase by 10\%, 15\%,
and 10\%, respectively, while the normalized
Use Spoon stage-completion score increases by 13.3\%.
For VLAct, the corresponding gains are 10\%, 5\%, and 5\%
in binary success and 6.7\% in stage completion.
Both augmented policies succeed in all 20 Press-button trials
and in 19 of 20 Carrot-in-pot trials.

The average task metric increases from 66.67\% to 78.75\%
for QwenPI and from 86.25\% to 92.92\% for VLAct, with gains
computed from the displayed averages. Improvements therefore
extend to a VLA-pretrained backbone paired with a GR00T action
head. Within each comparison, the base-policy checkpoint
remains fixed, so the gains do not arise from additional
base-policy fine-tuning.

\subsection{Qualitative Analysis of Recovery Behavior}
\label{app:real_world_recovery_behavior}

\begin{figure}[!htbp]
    \centering
    \includegraphics[width=\linewidth,height=0.82\textheight,keepaspectratio]{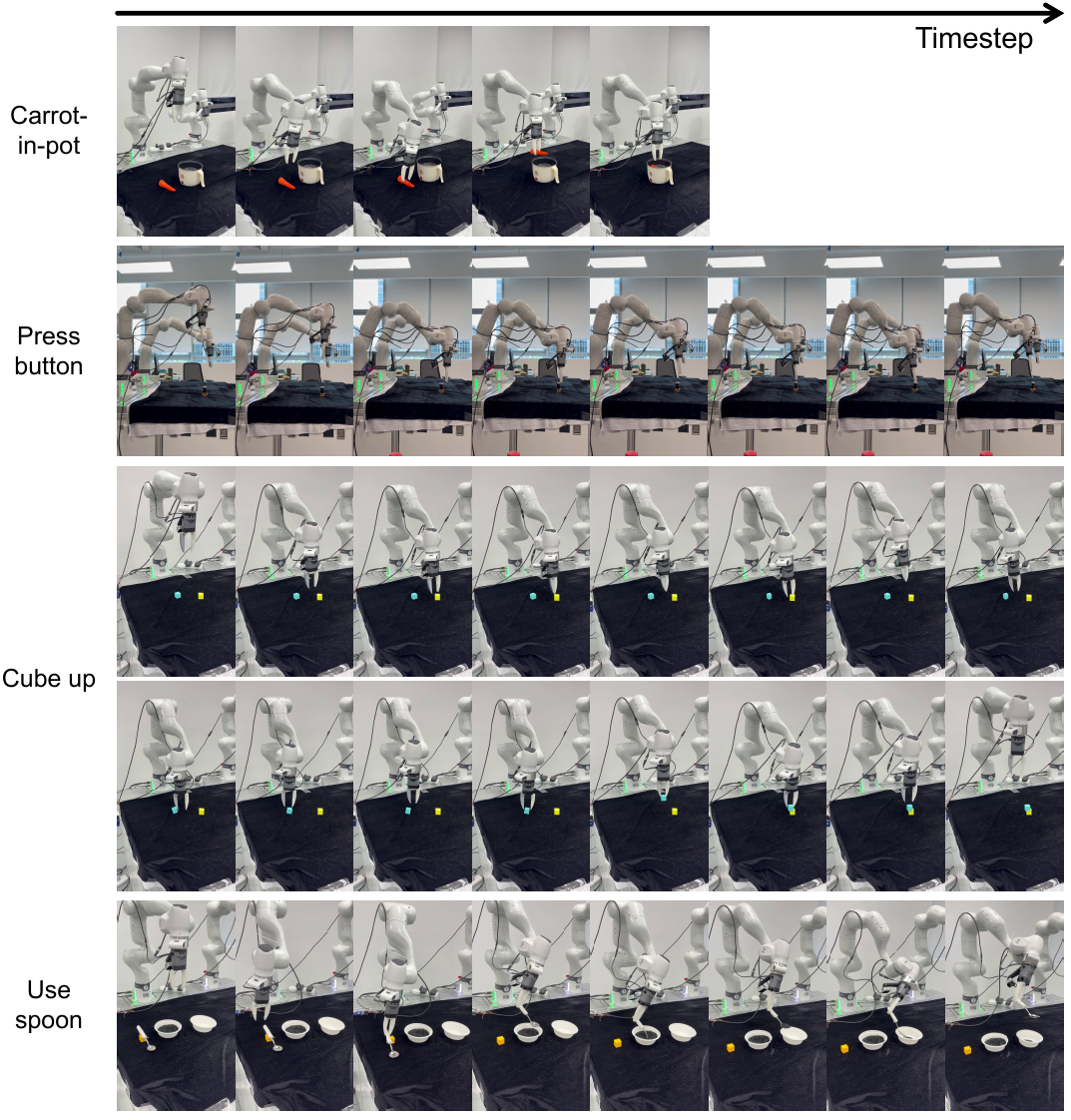}
    \caption{\textbf{Real-world execution with QwenPI+RESETTLE.}
    Task groups show Carrot-in-pot, Press-button, Cube-up, and
    Use Spoon from top to bottom. Frames progress left to right;
    Cube-up continues from its upper row to its lower row within
    one rollout. All sequences use QwenPI+RESETTLE, and individual
    recovery actions are not annotated. The selected frames show
    the combined system's execution, including repeated approaches
    before cube acquisition and subsequent stacking.}
    \label{fig:real_world_inference}
\end{figure}

Figure~\ref{fig:real_world_inference} shows QwenPI+RESETTLE
executions on all four tasks. Every sequence uses our method;
the two Cube-up rows continue the same rollout. Since individual
interventions are not marked, we analyze the overall behavior
of the combined system without assigning particular frames
to recovery or nominal policy control.

\paragraph{Object placement and contact manipulation.}
In Carrot-in-pot, the gripper approaches the carrot, establishes
a grasp, and carries it over the pot before placement. The
Press-button sequence shows changes in the approach pose and
repeated positioning near the button before withdrawal.
These sequences illustrate continued execution through object
acquisition and localized contact. The relevant correction
depends on the interaction: placement requires a usable grasp
and alignment with the receptacle, whereas pressing requires
alignment with the contact target.

\paragraph{Revisiting acquisition before stacking.}
The longer Cube-up sequence shows approaches that initially
do not lift a cube, followed by renewed positioning and
grasping. Continuing into the second row, the robot acquires
a cube, moves it over the other cube, and completes the stack.
The overall behavior preserves the physical prerequisite
for stacking: acquisition is achieved before the placement
motion completes. This case illustrates how execution can
return to an unfinished interaction and subsequently make
progress within the same rollout.

\paragraph{Maintaining progress through tool use.}
In Use Spoon, the sequence proceeds from approaching and
grasping the spoon to interacting with the source bowl and
moving toward the target bowl. The relevant manipulation
conditions change across these stages: the robot must first
establish a tool grasp, then position the spoon for scooping
and transfer. The frames show this progression, but do not
alone establish whether a bean was successfully transferred;
that outcome is assessed by the stage-completion metric.

\paragraph{Recovery modes and interpretation.}
Together, the sequences illustrate the roles of approach-pose
adjustment, renewed acquisition, and continuation after a
necessary manipulation condition has been established. These
behaviors are compatible with reference-guided local correction
followed by resumed base-policy execution, without an explicit
task-stage planner. In separate baseline observations on
Cube-up and Use Spoon, QwenPI sometimes made two unsuccessful
grasp attempts before proceeding to placement without holding
the object. Those observations motivate revisiting unfinished
stages; the present figure shows only QwenPI+RESETTLE rollouts
and is not a paired baseline comparison. The task-level results
in Table~\ref{tab:real_world_results} quantify the benefit of
adding recovery while keeping base-policy weights fixed.

\section{Per-Task Simulation Results}
\label{app:per_task_results}

We report per-task success rates (ASR, \%) for Meta-World,
RoboCasa Tabletop with GR1, and LIBERO-Pro Swap. ``Base.''
denotes the base configuration without recovery, and ``Ours''
denotes the same configuration augmented with RESETTLE.
Bold marks the better result within each Base./Ours pair,
including ties. Episode budgets and evaluation settings follow
Appendix~\ref{app:simulation_evaluation}.

\subsection{Meta-World MT50}
\label{app:metaworld_per_task}

Tables~\ref{tab:metaworld_per_task_00_24}
and~\ref{tab:metaworld_per_task_25_49} report QwenPI and
SmolVLA results on all 50 MT50 tasks, with 10 episodes per
task and configuration. Task IDs follow the evaluation order;
all task names have the \texttt{-v3} suffix, omitted in the
tables for compactness. These tables report only per-task
ASR. The aggregate evaluation remains the equally weighted
mean of the four difficulty-group success rates reported in
Table~\ref{tab:metaworld_difficulty_eval}.

\begin{table}[!htbp]
    \centering
    \caption{\textbf{Per-task ASR on Meta-World (tasks 00--24).} Success rates (\%) over 10 episodes per task. Ours denotes RESETTLE; bold marks the better result within each model pair, including ties.}
    \label{tab:metaworld_per_task_00_24}
    \vspace{5pt}
    \begin{tabular*}{\textwidth}{@{\extracolsep{\fill}}llcccc@{}}
        \toprule
        ID & Task & \multicolumn{2}{c}{QwenPI}
        & \multicolumn{2}{c}{SmolVLA} \\
        \cmidrule(lr){3-4} \cmidrule(lr){5-6}
        & & Base. & Ours & Base. & Ours \\
        \midrule
        00 & nut-assembly & \textbf{40} & \textbf{40} & \textbf{90} & \textbf{90} \\
        01 & basketball & \textbf{30} & \textbf{30} & \textbf{10} & \textbf{10} \\
        02 & bin-picking & \textbf{40} & \textbf{40} & 10 & \textbf{20} \\
        03 & box-close & 30 & \textbf{70} & 40 & \textbf{50} \\
        04 & button-press-topdown & \textbf{100} & \textbf{100} & \textbf{100} & \textbf{100} \\
        05 & button-press-topdown-wall & \textbf{100} & \textbf{100} & \textbf{100} & \textbf{100} \\
        06 & button-press & \textbf{100} & \textbf{100} & \textbf{100} & \textbf{100} \\
        07 & button-press-wall & \textbf{100} & \textbf{100} & \textbf{90} & \textbf{90} \\
        08 & coffee-button & \textbf{100} & \textbf{100} & \textbf{100} & \textbf{100} \\
        09 & coffee-pull & \textbf{80} & \textbf{80} & 10 & \textbf{20} \\
        10 & coffee-push & \textbf{90} & \textbf{90} & \textbf{90} & \textbf{90} \\
        11 & dial-turn & \textbf{0} & \textbf{0} & \textbf{60} & \textbf{60} \\
        12 & nut-disassemble & 60 & \textbf{80} & \textbf{0} & \textbf{0} \\
        13 & door-close & \textbf{100} & 90 & \textbf{100} & \textbf{100} \\
        14 & door-lock & \textbf{100} & \textbf{100} & \textbf{80} & 70 \\
        15 & door & \textbf{100} & \textbf{100} & \textbf{100} & \textbf{100} \\
        16 & door-unlock & \textbf{100} & \textbf{100} & \textbf{100} & \textbf{100} \\
        17 & hand-insert & \textbf{90} & \textbf{90} & 80 & \textbf{100} \\
        18 & drawer-close & \textbf{100} & \textbf{100} & \textbf{100} & 90 \\
        19 & drawer-open & \textbf{100} & \textbf{100} & 50 & \textbf{100} \\
        20 & faucet-open & \textbf{100} & \textbf{100} & 90 & \textbf{100} \\
        21 & faucet-close & \textbf{90} & \textbf{90} & \textbf{100} & \textbf{100} \\
        22 & hammer & \textbf{50} & \textbf{50} & 80 & \textbf{100} \\
        23 & handle-press-side & \textbf{100} & \textbf{100} & \textbf{100} & \textbf{100} \\
        24 & handle-press & \textbf{100} & \textbf{100} & \textbf{100} & \textbf{100} \\
        \bottomrule
    \end{tabular*}
\end{table}

\begin{table}[!htbp]
    \centering
    \caption{\textbf{Per-task ASR on Meta-World (tasks 25--49).} Success rates (\%) over 10 episodes per task. Ours denotes RESETTLE; bold marks the better result within each model pair, including ties.}
    \label{tab:metaworld_per_task_25_49}
    \vspace{5pt}
    \begin{tabular*}{\textwidth}{@{\extracolsep{\fill}}llcccc@{}}
        \toprule
        ID & Task & \multicolumn{2}{c}{QwenPI}
        & \multicolumn{2}{c}{SmolVLA} \\
        \cmidrule(lr){3-4} \cmidrule(lr){5-6}
        & & Base. & Ours & Base. & Ours \\
        \midrule
        25 & handle-pull-side & 10 & \textbf{20} & 20 & \textbf{30} \\
        26 & handle-pull & 10 & \textbf{30} & \textbf{100} & \textbf{100} \\
        27 & lever-pull & \textbf{40} & \textbf{40} & 0 & \textbf{10} \\
        28 & pick-place-wall & \textbf{70} & \textbf{70} & \textbf{40} & 20 \\
        29 & pick-out-of-hole & 0 & \textbf{30} & \textbf{40} & 30 \\
        30 & pick-place & \textbf{10} & \textbf{10} & \textbf{90} & \textbf{90} \\
        31 & plate-slide & \textbf{90} & \textbf{90} & \textbf{100} & 90 \\
        32 & plate-slide-side & \textbf{100} & \textbf{100} & \textbf{100} & 70 \\
        33 & plate-slide-back & \textbf{100} & \textbf{100} & 40 & \textbf{100} \\
        34 & plate-slide-back-side & 60 & \textbf{100} & \textbf{100} & 80 \\
        35 & peg-insertion-side & \textbf{40} & \textbf{40} & 0 & \textbf{100} \\
        36 & peg-unplug-side & 50 & \textbf{70} & 10 & \textbf{100} \\
        37 & soccer & 60 & \textbf{70} & \textbf{70} & \textbf{70} \\
        38 & stick-push & \textbf{70} & \textbf{70} & \textbf{100} & \textbf{100} \\
        39 & stick-pull & 60 & \textbf{70} & 0 & \textbf{70} \\
        40 & push & \textbf{100} & \textbf{100} & \textbf{100} & 30 \\
        41 & push-wall & \textbf{100} & \textbf{100} & \textbf{80} & 50 \\
        42 & push-back & 30 & \textbf{40} & \textbf{100} & 30 \\
        43 & reach & 20 & \textbf{30} & 30 & \textbf{60} \\
        44 & reach-wall & \textbf{90} & \textbf{90} & \textbf{60} & 10 \\
        45 & shelf-place & \textbf{40} & \textbf{40} & 20 & \textbf{100} \\
        46 & sweep-into-goal & \textbf{90} & \textbf{90} & \textbf{70} & 60 \\
        47 & sweep & 20 & \textbf{30} & 0 & \textbf{10} \\
        48 & window-open & \textbf{90} & \textbf{90} & \textbf{100} & \textbf{100} \\
        49 & window-close & \textbf{100} & \textbf{100} & \textbf{100} & \textbf{100} \\
        \bottomrule
    \end{tabular*}
\end{table}

Task-level effects vary across policies: both gains and
regressions occur, even though the difficulty-group aggregates
improve. These results do not imply uniform benefit from
recovery on every task.

\subsection{RoboCasa Tabletop / GR1}
\label{app:robocasa_per_task}

Table~\ref{tab:robocasa_per_task} reports ASR over 50 episodes
per task and configuration. IDs 01--24 follow the evaluation
order. For tasks 01--06, labels abbreviate
\texttt{PnP}\emph{Object}\texttt{To}\emph{Destination}\texttt{Close};
``close'' denotes the final drawer, microwave, or cabinet
closure. For tasks 07--24, the identifiers follow
\texttt{PosttrainPnPNovelFrom}\emph{Source}\texttt{To}\emph{Destination}\texttt{SplitA},
with source and destination tokens shown in the table.
The average weights all 24 tasks equally, matching the pooled
success rate because each task has the same episode budget.

\begin{table}[!htbp]
    \centering
    \caption{\textbf{Per-task ASR on RoboCasa Tabletop / GR1.} Success rates (\%) over 50 episodes per task. Ours denotes RESETTLE; bold marks the better result within each model pair, including ties.}
    \label{tab:robocasa_per_task}
    \vspace{5pt}
    \begin{tabular*}{\textwidth}{@{\extracolsep{\fill}}llcccc@{}}
        \toprule
        ID & Task & \multicolumn{2}{c}{QwenGR00T}
        & \multicolumn{2}{c}{LDA} \\
        \cmidrule(lr){3-4} \cmidrule(lr){5-6}
        & & Base. & Ours & Base. & Ours \\
        \midrule
        01 & Cup $\rightarrow$ Drawer (close) & \textbf{36} & 34 & \textbf{50} & 42 \\
        02 & Potato $\rightarrow$ Microwave (close) & 30 & \textbf{42} & \textbf{42} & 36 \\
        03 & Milk $\rightarrow$ Microwave (close) & \textbf{60} & 56 & 40 & \textbf{50} \\
        04 & Bottle $\rightarrow$ Cabinet (close) & \textbf{78} & 76 & 70 & \textbf{78} \\
        05 & Wine $\rightarrow$ Cabinet (close) & 40 & \textbf{58} & \textbf{72} & 58 \\
        06 & Can $\rightarrow$ Drawer (close) & 74 & \textbf{78} & \textbf{86} & 82 \\
        07 & Cuttingboard $\rightarrow$ Basket & 52 & \textbf{72} & 38 & \textbf{60} \\
        08 & Cuttingboard $\rightarrow$ Cardboardbox & \textbf{60} & \textbf{60} & 34 & \textbf{48} \\
        09 & Cuttingboard $\rightarrow$ Pan & 72 & \textbf{80} & \textbf{62} & \textbf{62} \\
        10 & Cuttingboard $\rightarrow$ Pot & 62 & \textbf{68} & \textbf{54} & \textbf{54} \\
        11 & Cuttingboard $\rightarrow$ Tieredbasket & 40 & \textbf{54} & 48 & \textbf{50} \\
        12 & Placemat $\rightarrow$ Basket & 34 & \textbf{58} & \textbf{50} & 44 \\
        13 & Placemat $\rightarrow$ Bowl & 64 & \textbf{76} & \textbf{44} & 40 \\
        14 & Placemat $\rightarrow$ Plate & 62 & \textbf{64} & 52 & \textbf{58} \\
        15 & Placemat $\rightarrow$ Tieredshelf & \textbf{22} & \textbf{22} & \textbf{34} & \textbf{34} \\
        16 & Plate $\rightarrow$ Bowl & \textbf{60} & 56 & 40 & \textbf{48} \\
        17 & Plate $\rightarrow$ Cardboardbox & 44 & \textbf{56} & \textbf{46} & \textbf{46} \\
        18 & Plate $\rightarrow$ Pan & 66 & \textbf{78} & 48 & \textbf{60} \\
        19 & Plate $\rightarrow$ Plate & 76 & \textbf{78} & 60 & \textbf{72} \\
        20 & Tray $\rightarrow$ Cardboardbox & 54 & \textbf{60} & \textbf{50} & 36 \\
        21 & Tray $\rightarrow$ Plate & 64 & \textbf{66} & 60 & \textbf{66} \\
        22 & Tray $\rightarrow$ Pot & 56 & \textbf{68} & 44 & \textbf{46} \\
        23 & Tray $\rightarrow$ Tieredbasket & \textbf{54} & 52 & 58 & \textbf{60} \\
        24 & Tray $\rightarrow$ Tieredshelf & 28 & \textbf{40} & 26 & \textbf{28} \\
        \midrule
        \multicolumn{2}{l}{Average} & 53.67 & \textbf{60.50} & 50.33 & \textbf{52.42} \\
        \bottomrule
    \end{tabular*}
\end{table}

The averages reproduce Table~\ref{tab:robocasa_tabletop}:
53.67\% to 60.50\% for QwenGR00T and 50.33\% to 52.42\%
for LDA. Task-level gains coexist with regressions and ties,
so the aggregate improvements reflect differing responses
across manipulation tasks.

\subsection{LIBERO-Pro Swap}
\label{app:libero_pro_per_task}

Table~\ref{tab:libero_pro_per_task} reports the 10 tasks in
\texttt{libero\_10\_swap}, using seeds 0--9 for 10 episodes
per task and configuration. Harness VLA denotes the
Codex/GPT-5.5 planner route with $\pi_{0.5}$ as the low-level
VLA. The direct $\pi_{0.5}$ route does not use that planner.
Each route is evaluated with and without RESETTLE.

\begin{table}[!htbp]
    \centering
    \caption{\textbf{Per-task ASR on LIBERO-Pro Swap.} Success rates (\%) over 10 episodes per task. Ours denotes RESETTLE; bold marks the better result within each route, including ties.}
    \label{tab:libero_pro_per_task}
    \vspace{5pt}
    \begin{tabular*}{\textwidth}{@{\extracolsep{\fill}}lcccc@{}}
        \toprule
        Task & \multicolumn{2}{c}{Direct $\pi_{0.5}$}
        & \multicolumn{2}{c}{Harness VLA} \\
        \cmidrule(lr){2-3} \cmidrule(lr){4-5}
        & Base. & Ours & Base. & Ours \\
        \midrule
        0 & 0 & \textbf{10} & 70 & \textbf{80} \\
        1 & 40 & \textbf{50} & 60 & \textbf{70} \\
        2 & 10 & \textbf{20} & \textbf{50} & \textbf{50} \\
        3 & \textbf{0} & \textbf{0} & \textbf{0} & \textbf{0} \\
        4 & 30 & \textbf{70} & 50 & \textbf{80} \\
        5 & 60 & \textbf{70} & \textbf{80} & \textbf{80} \\
        6 & \textbf{10} & \textbf{10} & 30 & \textbf{40} \\
        7 & 0 & \textbf{10} & 70 & \textbf{80} \\
        8 & \textbf{0} & \textbf{0} & 10 & \textbf{20} \\
        9 & \textbf{0} & \textbf{0} & \textbf{0} & \textbf{0} \\
        \midrule
        Average & 15.00 & \textbf{24.00} & 42.00 & \textbf{50.00} \\
        \bottomrule
    \end{tabular*}
\end{table}

RESETTLE improves or matches the baseline on every task in
both routes, while some tasks remain unsolved. The averages
match Table~\ref{tab:libero_pro_swap}: 15\% to 24\% for direct
$\pi_{0.5}$ execution and 42\% to 50\% for Harness VLA.

\end{document}